\documentclass[10pt,journal,compsoc]{IEEEtran}

\ifCLASSOPTIONcompsoc
  \usepackage[nocompress]{cite}
\else
  \usepackage{cite}
\fi

\usepackage{xspace} %
\usepackage[pdftex]{graphicx}

\usepackage{ifthen}  %
\usepackage{subcaption} %
\usepackage{amsmath}
\usepackage{mathtools}
\usepackage{booktabs}
\usepackage[normalem]{ulem}
\usepackage{amssymb}
\usepackage[dvipsnames]{xcolor}  %
\AtBeginDocument{\colorlet{defaultcolor}{.}}
\usepackage{pdfpages} %

\usepackage{amsbsy} 
\usepackage{array}  %
\usepackage{arydshln}
\usepackage[hyphens]{url}
\usepackage{multirow}
\usepackage{enumitem} %
\usepackage[export]{adjustbox} %
\usepackage[hidelinks]{hyperref}
\usepackage{hyperref}
\definecolor{BlueBlack}{RGB}{36, 113, 163}
\hypersetup{
    colorlinks,
    linkcolor={red},
    citecolor={BlueBlack},
    urlcolor={blue}
}
\usepackage{cleveref}[capitalise]
\crefname{section}{Sec.}{Secs.}
\crefname{figure}{Fig.}{Figs.}
\crefname{table}{Tab.}{Tabs.}
\crefname{equation}{Eq.}{Eqs.}
\Crefname{section}{Section}{Sections}
\usepackage{url}

\ExplSyntaxOn
\cs_new:Npn \expandableinput #1
  { \use:c { @@input } { \file_full_name:n {#1} } }
\AddToHook{env/tabular/begin}
  { \cs_set_eq:NN \input \expandableinput }
\ExplSyntaxOff

\newcommand{\bcos}{B-cos\xspace}
\newcommand{\mbcos}{mB-cos\xspace}

\newcommand{\bcosnets}{\bcos Networks\xspace}
\newcommand{\imnet}{ImageNet\xspace}
\newcommand{\bcosnet}{\bcos Network\xspace}
\newcommand{\gcam}{GradCAM\xspace}

\makeatletter
\newcommand{\ie}{\textit{i}.\textit{e}.\@ifnextchar{,}{}{~}}
\newcommand{\Ie}{\textit{I}.\textit{e}.\@ifnextchar{,}{}{~}}
\makeatother
\makeatletter
\newcommand{\eg}{\textit{e}.\textit{g}.\@ifnextchar{,}{}{~}}
\newcommand{\Eg}{\textit{E}.\textit{g}.\@ifnextchar{,}{}{~}}
\makeatother

\newboolean{showcomments}
\setboolean{showcomments}{false}
\newboolean{showrev}
\setboolean{showrev}{true}

\ifthenelse{\boolean{showcomments}}{
    \newcommand{\mario}[1]{\textbf{\textcolor{blue}{Mario:\ #1}}}
    \newcommand{\moritz}[1]{\textbf{\textcolor{orange}{Moritz:\ #1}}}
    \newcommand{\singh}[1]{{\color[RGB]{0, 95, 0}#1}}

    \newcommand{\bernt}[1]{\textcolor[rgb]{0.82, 0.1, 0.26}{\textbf{Bernt: #1}}}
    }
{
    \newcommand{\mario}[1]{}
    \newcommand{\moritz}[1]{}

    \newcommand{\bernt}[1]{}
    \newcommand{\singh}[1]{}
}
\ifthenelse{\boolean{showrev}}{
    
    }
{
       
}
\makeatletter
\def\thickhline{%
  \noalign{\ifnum0=`}\fi\hrule \@height \thickarrayrulewidth \futurelet
   \reserved@a\@xthickhline}
\def\@xthickhline{\ifx\reserved@a\thickhline
               \vskip\doublerulesep
               \vskip-\thickarrayrulewidth
             \fi
      \ifnum0=`{\fi}}
\makeatother

\newlength{\thickarrayrulewidth}
\newcommand{\myparagraph}[2][.5]{\vspace{#1em}\noindent{\bf #2}}

\newcommand{\mat}[1]{\MakeUppercase{\mathbf{#1}}}
\renewcommand{\vec}[1]{\MakeLowercase{\mathbf{#1}}}
\newcommand{\inputset}[1]{\boldsymbol{\mathcal{#1}}}

\newcommand\myeq{\mkern1.25mu{=}\mkern1.25mu}

\newcommand\myminus{\mkern1.25mu{-}\mkern1.25mu}
\newcommand\myplus{\mkern1.25mu{+}\mkern1.25mu}
\newcommand\mypropto{\mkern1.25mu{\propto}\mkern1.25mu}
\newcommand\mygreater{\mkern1.25mu{>}\mkern1.25mu}

\newcommand\myin{\mkern1.25mu{\in}\mkern1.25mu}
\newcommand\mytimes{\mkern1.25mu{\times}\mkern1.25mu}

\begin{document}

\title{
\bcos Alignment for Inherently Interpretable\\ CNNs and Vision Transformers

}
\markboth{Preprint. Accepted for Publication in IEEE Transactions on Pattern Analysis and Machine Intelligence.}%
{Böhle \MakeLowercase{\textit{et al.}}: \mytitle}

\author{Moritz~Böhle,
        Navdeeppal~Singh,
        Mario~Fritz, and~Bernt~Schiele~\IEEEmembership{Fellow,~IEEE}
\IEEEcompsocitemizethanks{\IEEEcompsocthanksitem Moritz Böhle is the corresponding author. Moritz Böhle, Navdeeppal Singh, and Bernt Schiele are with the Department of Computer Vision and Machine Learning, Max Planck Institute for Informatics, Saarland Informatics Campus, Saarbrücken, 66123, Germany. E-mail: \{mboehle,nsingh,schiele\}@mpi-inf.mpg.de
\protect\\
\IEEEcompsocthanksitem
Mario Fritz is with the CISPA Helmholtz Center for Information Security, Saarbrücken 66123, Germany. E-mail: fritz@cispa.de.
}
\thanks{
}
}

\IEEEtitleabstractindextext{%
\begin{abstract}
We present a new direction for increasing the interpretability of deep neural networks (DNNs) by promoting weight-input alignment during training. 
For this, we propose to replace the linear transformations in DNNs by our novel \bcos transformation. As we show, a sequence (network) of such transformations induces a single linear transformation that faithfully summarises the full model computations. Moreover, the \bcos transformation is designed such that the weights align with relevant signals during optimisation. As a result, those induced linear transformations become highly interpretable and highlight task-relevant features. Importantly, the \bcos{} transformation is designed to be compatible with existing architectures and we show that it can easily be integrated into virtually all of the latest state of the art models for computer vision---\eg ResNets, DenseNets, ConvNext models, as well as Vision Transformers---by combining the B-cos-based explanations with normalisation and attention layers, all whilst maintaining similar accuracy on \imnet. Finally, we show that the resulting explanations are of high visual quality and perform well under quantitative interpretability metrics.  
\end{abstract}

\begin{IEEEkeywords}
Interpretable Deep Learning, XAI, Convolutional Neural Networks, Vision Transformers, Interpretability
\end{IEEEkeywords}}

\maketitle
\IEEEdisplaynontitleabstractindextext
\IEEEpeerreviewmaketitle

\IEEEraisesectionheading{\section{Introduction}\label{sec:intro}}

\IEEEPARstart{W}{hile} deep neural networks (DNNs) are highly successful in a wide range of tasks, explaining their decisions remains an open research problem~\cite{xai_overview}.
The difficulty here lies in the fact that such explanations need to faithfully summarise the internal model computations \emph{and} present them in a {human-interpretable} manner. 
\Eg, it is well known that piece-wise linear (\ie ReLU-based \cite{nair2010relu}) models  are accurately summarised by a linear transformation for every input~\cite{montufar2014number}.
However, despite providing an accurate summary, these piece-wise linear transformations are generally
not intuitively interpretable for humans and tend to perform poorly under quantitative interpretability metrics, cf.~\cite{BAM2019, shrikumar2017deeplift}. 
Recent work thus aimed to improve the explanations' interpretability, often focusing on their visual quality~\cite{adebayo2018sanity}. However, gains in the visual quality of the explanations could come at the cost of their model-faithfulness~\cite{adebayo2018sanity}.

\begin{figure*}[!ht]
    \centering
    \includegraphics[width=\textwidth]{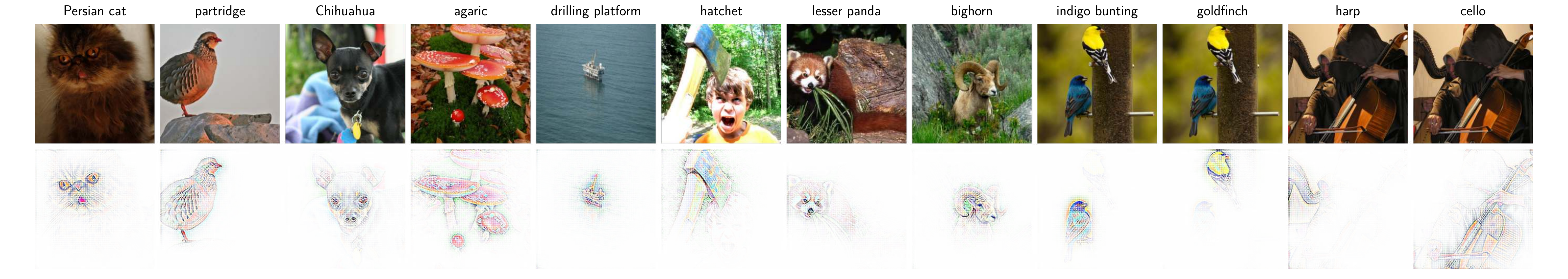}
    \vspace{-.5em}
    \caption{\textbf{Top:} Inputs $\vec x_i$ to a \bcos{} DenseNet-121. \textbf{Bottom:} \bcos{} explanation for class $c$ ($c$: image label). Specifically, we visualise the $c$-th row of $\mat w_{1\rightarrow L}(\vec x_i)$ as applied by the model, see \cref{eq:collapse}; {no masking of the original image is used}. For the last 2 images, we also show the explanation for the 2nd most likely class. For details on visualising $\mat w_{1\rightarrow L}(\vec x_i)$, see \cref{sec:experiments}. \newline}
    \label{fig:global_protos}
    \vspace{-1em}
\end{figure*}

Instead of optimising the explanation method, we aim to design the DNNs to inherently provide an explanation that fulfills the aforementioned requirements---the resulting explanations constitute both a faithful summary and have a clear interpretation for humans.
To achieve this, we propose the \textbf{\bcos transformation} as a drop-in replacement for linear transformations. 
As such, the \bcos transformation can easily be integrated into a wide range of existing DNN architectures and we show that the resulting models provide high-quality explanations for their decisions, see \cref{fig:global_protos}.

To ensure that these explanations constitute a faithful summary of the models, we design the \bcos transformation as an input-dependent linear transformation. 
Importantly, any sequence of such transformations therefore induces a 
single linear transformation that faithfully summarises the entire sequence. To make the induced linear transformations interpretable, the \bcos transformation is designed such that its weights align with task-relevant input signals during optimisation. The summarising linear transformation thus becomes easily interpretable for humans:
it is a direct reflection of the weights the model has learnt during training and specifically reflects those weights that best align with a given input.  In short, we make the following contributions:

\begin{enumerate}[wide, label={\textbf{(\arabic*)}}, itemsep=0em, topsep=-1.25em, labelwidth=0em, labelindent=0pt]
    \item We introduce the \bcos transformation to improve the interpretability of DNNs. By promoting weight-input alignment, these transformations are explicitly designed to yield explanations that highlight task-relevant input patterns.
    \item Specifically, the \bcos transformation is designed such that any sequence of \bcos transformations can be {faithfully summarised} by a single linear transformation.
    We show that this allows to explain not only the models' outputs, but also representations in intermediate network layers. 
    \item 
    We demonstrate that a plain \bcos convolutional neural network without any additional non-linearities, normalisation layers or regularisation schemes achieves competitive performance on  CIFAR10~\cite{krizhevsky2009cifar10}, demonstrating the modelling capacity of DNNs that are solely based on the \bcos transformation. 
    In this context, we show that the parameter B gives fine-grained control over the increase in weight alignment and thus the interpretability of the \bcos networks.
    {\item 
    We analyse how to integrate normalisation layers into \bcos models to take advantage of the optimisation benefits of those layers without sacrificing interpretability. %
    \item 
    We analyse how to combine the proposed \bcos framework with attention-based models (\eg Vision Transformers (ViTs) \cite{dosovitskiy2021an}), which highlights the generality of our approach and shows that it extends beyond convolutional networks.
    \item 
    Finally, we show in a wide range of experiments that \bcos DNNs achieve similar accuracy as their conventional counterparts. Specifically, we evaluate the \bcos versions of various commonly used DNNs such as VGGs~\cite{simonyan2015vgg}, %
    ResNets~\cite{he2016deep}, DenseNets~\cite{huang2017densely}, ResNeXt~\cite{xie2017aggregated} and ConvNeXt models \cite{liu2022convnet}, as well as  ViTs \cite{dosovitskiy2021an}. %
    Our strongest models achieve $>$80\% top-1 accuracy on ImageNet, all while providing highly detailed explanations for their decisions. Specifically, the \bcos explanations outperform other explanation methods across all tested architectures, both under quantitative metrics as well as under qualitative inspection.}
\end{enumerate}\vspace{1.25em}

{As such, this work extends \cite{Boehle2022CVPR} by integrating normalisation and attention layers into \bcos DNNs and provides a significantly broader experimental evaluation of the proposed framework. The code and all of the trained models are  publicly available at \href{https://www.github.com/B-cos/B-cos-v2}{github.com/B-cos/B-cos-v2}.}

\section{Related work}
\label{sec:related}
\myparagraph[-.25]{Approaches for understanding DNNs}
 typically focus on explaining individual model decisions \emph{post-hoc}, i.e., they are designed to work on any pre-trained DNN. Examples of this include perturbation-based, \cite{lundberg2017unified,petsiuk2018rise,ribeiro2016lime}, activation-based, \cite{kim2018tcav,das2020prototypes}, or backpropagation-based explanations, \cite{simonyan2013deep,springenberg2014striving,zhou2016CAM,bach2015pixel,selvaraju2017grad,shrikumar2017deeplift,srinivas2019full,sundararajan2017axiomatic}. 
 In order to obtain explanations for the \bcos networks, we also rely on a backpropagation-based approach. In contrast to post-hoc explanation methods, however, {we design the \bcos networks to be explainable under this particular form of backpropagation and the resulting explanations are thus {model-inherent.}}
 
 The design of such \emph{inherently interpretable} models has gained increased attention recently. Examples include prototype-based networks~\cite{chen2019looks}, BagNets~\cite{brendel2018approximating}, and CoDA Nets~\cite{Boehle2021CVPR}. 
 Similar to the BagNets and the CoDA Nets, our \bcos networks can be faithfully summarised by a single linear transformation. 
 Moreover, similar to~\cite{Boehle2021CVPR}, we rely on a structurally induced alignment pressure to make those transformations interpretable.
 In contrast to those works, however, our method is specifically designed to be compatible with existing neural network architectures, which allows us to improve the interpretability of a wide range of DNNs.

\myparagraph{Weight-input alignment.} It has been observed that adversarial training promotes alignment~\cite{tsipras2019atodds} and recent studies suggest that this could increase interpretability via gradient-based explanations~\cite{Shah2021grads,kim2019safeml}. Further, \cite{srinivas2021rethink} introduce a loss to increase alignment. Instead of relying on loss-based model regularisation, the increase in alignment in \bcos networks is based on \emph{architectural constraints} that promote weight-input alignment during model optimisation. {Finally, \cite{kindermans2018learning} note that the weights of a (piece-wise) linear classifier cannot be expected to align with relevant input signals in the presence of distractors (\eg the background in object classification datasets). To overcome this, \cite{kindermans2018learning} propose to learn the relevant input signals for a given pre-trained model from data in order to explain said model. In contrast, by optimising the weights to align with the relevant signals already during training, for \bcos models the weights themselves can directly be used to visualise the relevant input signals. }

\myparagraph{Non-linear transformations.}
While the linear transformation is the default operation for most neural network architectures, many non-linear transformations have been investigated \cite{zoumpourlis2017non,liu2017hyperspherical,Liu2018CVPR,luo2018cosine,ghiasi2019generalizing,wang2019kervolutional}. Most similar to our work are~\cite{liu2017hyperspherical,Liu2018CVPR,luo2018cosine}, which assess transformations that emphasise the cosine similarity (i.e., `alignment') between weights and inputs to improve model performance. In fact, we found that amongst other transformations, \cite{Liu2018CVPR} evaluates a non-linear transformation that is equivalent to our \bcos operator with $\text{B}\myeq2$ {as introduced in \cref{eq:bcos}}. 
{In contrast to \cite{Liu2018CVPR}, we explicitly introduce this non-linear transformation to increase model interpretability and show that the resulting models scale to large-scale classification problems.}

{
\myparagraph{The effect of biases on performance and interpretability.}
While `Input$\mytimes$Gradient' (IxG) \cite{shrikumar2017deeplift} accurately summarises a piece-wise \emph{linear} model given by\footnote{
    Here, we write $\mat W(\vec x) \myeq \mat W_{\vec x}$ to emphasise the fact that piece-wise linear models compute a linear transformation of the input. Thus, the gradient with respect to the input is given by the linear matrix $\mat W_{\vec x}$.
} $\vec y(\vec x) \myeq \mat W_{\vec x}^T\vec x$, most piece-wise linear models in fact additionally contain bias terms and compute $\vec y(\vec x) \myeq \mat W_{\vec x}^T\vec x + \vec b(\vec x)$. 
As bias terms are not accounted for in the explanations $\mat W_{\vec x}$, they do not provide a \emph{complete} description of the model~\cite{wang2019bias,srinivas2019full}. 
To overcome this \cite{wang2019bias,srinivas2019full} propose to adapt the explanation methods to reflect the biases  in the explanations, while \cite{mohan2019robust,hesse2021fast} show that, for some tasks, removing the bias terms altogether only has a minor effect on the model performance but can significantly improve the models' interpretability.
Similarly, we show that including bias terms in the models only provides minor improvements in model accuracy. Hence, we design the \bcos DNNs to be bias-free. This ensures that the linear model summaries constitute \emph{complete} explanations \cite{sundararajan2017axiomatic}.}

{
\myparagraph{Normalisation layers.} To design bias-free, yet easily optimisable and performant networks, we adapt the normalisation layers in \bcos DNNs such that they do not add biases to the model computations at inference time, similar to the modified BatchNorm in \cite{mohan2019robust}. 

\myparagraph{Explaining attention-based models.}
As the name exemplifies, attention is often thought to give insight into what a model `pays attention to' for its prediction. As such, various methods for using attention to understand the model output have been proposed, such as visualising the attention of single attention heads \cite{vaswani2017attention}. However, especially in deep layers the information can be highly distributed and it is unclear whether a given token still represents its original position in the input~\cite{serrano2019attention,abnar2020quantifying}, which complicates the interpretation of high attention scores deep in the neural network~\cite{serrano2019attention,bastings2020}. 

Therefore, \cite{abnar2020quantifying} proposed  `attention rollout', which summarises the various attention maps throughout the layers. However, this summary still only includes the attention layers and neglects all other network components~\cite{bastings2020}. In response, various improvements over attention rollout have been proposed, such as GradSAM~\cite{barkan2021GradSAM} or an LRP-based explanation method~\cite{chefer2021transformer}, that were designed to more accurately reflect the computations of \emph{all} model components. %

Instead of providing post-hoc explanations for ViT models, in this work we show that attention layers are inherently compatible with the \bcos-based explanations, as they also compute dynamic linear transformations of their inputs. As such, they can be combined seamlessly with \bcos layers to yield \bcos ViTs, which achieve similar accuracies as their conventional counterparts. As the resulting explanations inherently comprise the multi-layer perceptrons as well as the attention layers, they faithfully summarise the entire model and thus do not require any post-hoc explanations.
}
\section{\bcos neural networks}
\label{sec:method}

In this section, we introduce the \bcos transformation as a replacement for the linear units in DNNs, which are (almost) ``at the heart of every deep network''~\cite{swish}, and discuss how this can increase the interpretability of DNNs. 

For this, we first introduce the \bcos transformation as a variation of the linear transformation (\ref{subsec:bcos}) and highlight its most important properties.
Then, we show how to construct {\bcosnets} (\ref{subsec:bcos_net}) and how to {faithfully summarise} the network computations to obtain explanations for their outputs (\ref{subsubsec:explain}).
Subsequently, we discuss how the \bcos transformation---combined with the binary cross entropy (BCE) loss---affects the parameter optima of the models (\ref{subsubsec:optim}). 
Specifically, by inducing alignment pressure, the \bcos transformation aligns the model weights with task-relevant patterns in the input. Finally, in \cref{subsec:deep_bcos} we integrate the \bcos transformation into conventional DNNs by using it as a drop-in replacement for the ubiquitously used linear units and discuss how to combine \bcos layers with normalisation and attention layers without sacrificing the interpretability of \bcos networks.

\subsection{The \bcos transformation}
\label{subsec:bcos}

Typically, the individual `neurons' in a DNN compute the dot product between their weights $\vec w$ and an input $\vec x$:
\begin{align}
\label{eq:lc}
    f(\vec x; \vec w) &= \vec w^T\,\vec x = \lVert\vec w\rVert\, \lVert\vec x\rVert \, c(\vec x, \vec w)\;\text{,}\\
    \text{with}\quad c(\vec x, {\vec w})&= \cos\left(\angle(\vec x, {\vec w})\right) \; {.}
\end{align}
Here, $\angle(\vec x, {\vec w})$ returns the angle between the vectors $\vec x$ and ${\vec w}$.
As we seek to improve the interpretability of DNNs by promoting weight-input alignment during optimisation, we propose {to explicitly emphasise the alignment term in the linear transformation, \ie the impact of the cosine function. This yields the}
\textbf{\bcos transformation}:
\begin{align} 
    \label{eq:bcos}
    \hspace{-.07em}\text{\bcos} (\vec x; \vec w) 
    &\hspace{.075em}\myeq\underbrace{\lVert\widehat{\vec w} \rVert}_{\color[RGB]{10, 132, 180}=1}\hspace{.15em}\lVert\vec x \rVert\hspace{.15em} {\color[RGB]{10, 132, 180}\lvert}
    c(\vec x, \widehat{\vec w}){\color[RGB]{10, 132, 180}\rvert^{\text{B}} \mytimes \,\text{sgn}\left(c(\vec x, \widehat{\vec w})\right)}\,.
 \vspace{-.45em}\end{align}
{
Here, the hat-operator scales $\widehat{\vec w}$ to unit norm, B is a scalar, and sgn the sign function.
 The \bcos transformation thus rescales the output of a linear transformation and equals:
 \begin{align} 
    \label{eq:bcos2}
    \hspace{-.07em}\text{\bcos} (\vec x; \vec w) 
    &\hspace{.075em}\myeq {\color[RGB]{10, 132, 180}\widehat{\color{defaultcolor}\vec w}}^T \hspace{.05em}\vec x \hspace{.05em} {\color[RGB]{10, 132, 180}
    \times \lvert \cos(\vec x, \widehat{\vec w})\rvert^{\text{B-1}}}\,;
 \end{align}
 for a derivation of the equivalence, see supplement (Sec.~D).}
 
 Note that this only introduces \emph{minor changes} (highlighted in blue) with respect to~\cref{eq:lc}; \eg, for $\text{B}\myeq1$, \cref{eq:bcos2} is equivalent to a linear transformation with $\widehat{\vec w}$.
However, albeit small, these changes are important for {three reasons}. 
 
 \myparagraph{{First}}, they bound the output of \bcos neurons:
 \begin{align} 
\label{eq:bound}
    \lVert\widehat{\vec w}\rVert=1\; \Rightarrow \;\text{\bcos}(\vec x; \vec w) \leq \lVert\vec x\rVert \;.
\end{align}
As becomes clear from \cref{eq:bcos}, equality in \cref{eq:bound} is only achieved if $\vec x$ and $\vec w$ are collinear, i.e., \emph{aligned}. 

\begin{figure}
    \centering
    \begin{subfigure}[b]{\linewidth}
    \includegraphics[width=\linewidth, trim=1em 0 0 0, clip]{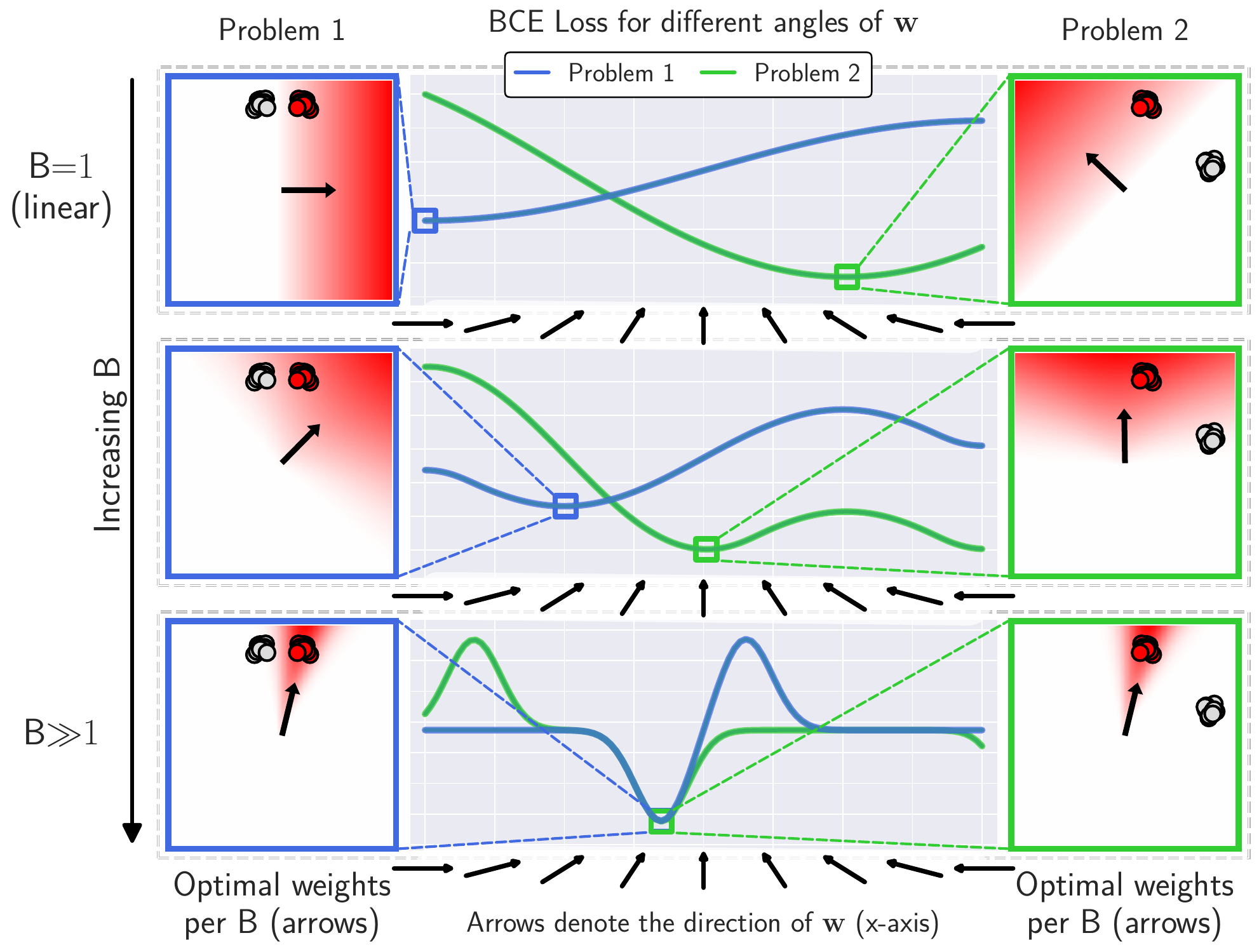}
    \end{subfigure}
    \caption{
    \textbf{Col.~2:} BCE loss for different angles of $\vec w$ for \bcos classifiers (\cref{eq:bcos}) with different values of B (rows) for two classification problems. \textbf{Cols.~1+3:} Visualisation of the classification problems and the corresponding optimal weights (arrows) per B. For $\text{B}\myeq1$ (first row) the weights $\vec w$ represent the decision boundary of a linear classifier. 
    Although the red cluster is the same in both cases, the optimal weight vectors differ significantly (compare within row). In contrast, for higher values of B the weights converge to the same optimum in both tasks (see last row). {The opacity of the red shading shows the strength of the positive activation of the B-cos transformation for a sample at a given position.}}
    \label{fig:toy_example}
\end{figure}

{
\myparagraph{{Secondly}}, {an increased exponent B further suppresses the outputs for weight-input pairs with low cosine similarity,} 
\begin{align}
\label{eq:supp}
    \text{B}\gg1 \land
    \lvert c(\vec x, \widehat{\vec w})\rvert <1 \,\Rightarrow\, \text{\bcos}(\vec x; {\vec w}) \ll \lVert\vec x\rVert\;,%
\end{align}
and the respective \bcos unit only yields outputs close to its maximum (\ie $\lVert\vec x\rVert$) for a small range of angular deviations from $\vec x$. Together, these two properties can significantly change the optimal parameters.
 To illustrate this, we show in \cref{fig:toy_example} how increasing B affects a simple linear classification problem. In particular, note how increasing B narrows the 'response window' (red area) of the \bcos transformation and suppresses the influence of the distractors (grey circles). In contrast to a linear classifier, which has a highly task-dependent optimum (cf.~first row in \cref{fig:toy_example}), the \bcos transformation thus classifies based on \emph{cosine similarity} between weights and inputs. As such, the optimal weights constitute 'angular prototypes' and lend themselves well for explaining the model prediction: a sample is confidently classified as the red class if it 'looks like' the weight vector.}

{
\myparagraph{Lastly}, the \bcos transformation maintains an important property of the linear transformation: specifically, sequences of \bcos transformations are faithfully summarised by a single linear transformation (\cref{eq:collapse}). As a result, the explanations for individual neurons in a DNN can easily be combined to yield explanations for the full model, as we discuss in \cref{subsubsec:explain}. Importantly, these explanations will align with discriminative patterns when optimising a \bcos network for classification (\cref{subsubsec:optim}), making them easily interpretable for humans.
}

\subsection{Simple (convolutional) \bcos networks}
\label{subsec:bcos_net}

In this section, we discuss how to construct simple (convolutional) DNNs based on the \bcos transformation. Then, we show how to summarise the network outputs by a single linear transformation and, finally, why this transformation aligns with discriminative patterns in classification tasks.

\myparagraph{\bcos networks.} The \bcos transformation is designed to replace the linear transformation and can be used in exactly the same way. \Eg, consider a \emph{conventional} fully connected multi-layer neural network
$\vec f(\vec x; \theta)$ of $L$ layers, defined as
\begin{align} 
\label{eq:relu_net}
    \vec f(\vec x; \theta) = \vec l_L   \circ\vec l_{L-1}   \circ...   \circ\vec l_{2}   \circ\vec l_1 (\vec x)\; \text{,}
\end{align}
with $\vec l_j$ denoting layer $j$ with parameters $\vec w^k_j$ for neuron $k$ in layer $j$, and $\theta$ the collection of all model parameters. In such a model, each layer $\vec l_j$ typically computes 
\begin{align} 
    \vec l_j(\vec a_j; \mat w_j) = \phi\left(\mat w_j \,\vec a_j\right)\;,
\end{align}
with $\vec a_j$ the input to layer $j$, $\phi$ a non-linear activation function (\eg, ReLU), and the row $k$ of $\mat w_j$ given by the weight vector $\vec w_j^k$ of the $k$-th neuron in that layer.
Note that the non-linear activation function $\phi$ is \emph{required} to be able to model non-linear relationships with multiple layers. \\
A corresponding \bcos network $\vec f^*$ with layers $\vec l^*_j$ can be formulated in exactly the same way as
\begin{align} 
\label{eq:bcos_net}
    \vec f^*(\vec x; \theta) =\vec l^*_L   \circ\vec l^*_{L-1}   \circ...   \circ\vec l^*_{2}   \circ\vec l^*_1 (\vec x)\,,
\end{align}
with the only difference being that every dot product (here between rows of $\mat w_j$ and inputs $\vec a_j$) is replaced by the \bcos transformation in \cref{eq:bcos2}.
In matrix form, this equates to
\begin{align}
    \vec l^*_j(\vec a_j; \mat w_j) =
         \left(\widehat{\mat w}_j \,\vec a_j \right)\times \lvert c(\vec a_j; \widehat{\mat w}_j)\rvert^{\text{B}-1}\label{eq:bcos_suppress}
        \; .
\end{align}
Here, the power, absolute value, and $\times$ operators are applied element-wise, $c(\vec a_j; \widehat{\mat w}_j)$ computes the cosine similarity between input $\vec a_j$ and the rows of $\widehat{\mat w}_j$, 
and the hat operator scales the rows of $\widehat{\mat w}_j$ to unit norm.
Finally, as for B$\mygreater1$ the transformation $\vec l_j^*$ is already \emph{non-linear}, {a non-linear $\phi$} is not required for modelling non-linear relationships.

The above discussion readily generalises to convolutional neural networks (CNNs): in CNNs, we replace the linear transformations computed by the convolutional kernels by \bcos, see Alg.\ 1 in appendix C. Further, although we assumed a plain multi-layer network without add-ons such as skip connections, we show in \cref{sec:results} that the benefits of \bcos also transfer to more advanced architectures (\cref{subsec:deep_bcos}).

\subsubsection{Computing explanations for \bcosnets}
\label{subsubsec:explain}

{As can be seen by rewriting \cref{eq:bcos_suppress}, a \bcos layer effectively computes an input-dependent linear transformation:
\begin{align} 
\label{eq:dynlin}
    \vec l^*_j(\vec a_j; \mat w_j) &\;=\; 
        \widetilde{\mat w}_j(\vec a_j) \,\vec a_j \, ,\\[.25em]
        \text{with} \quad {\widetilde{\mat w}_j(\vec a_j)}&\;=\;{|c(\vec a_j; \widehat{\mat w}_j)|^{\text{B}-1}\odot \widehat{\mat w}_j} \; \text{.}
        \label{eq:dynlin2}
 \end{align}
Here, $\odot$ scales the rows of the matrix to its right by the scalar entries of the vector to its left.
Hence, the output of a \bcosnet, see \cref{eq:bcos_net}, is effectively calculated as
\begin{align}
\label{eq:linear_sequence}
    \vec f^*(\vec x; \vec \theta) = \widetilde{\mat w}_L(\vec a_L)\widetilde{\mat w}_{L-1}(\vec a_{L-1})...\widetilde{\mat w}_1(\vec a_1\myeq\vec x)\vec x\;.
\end{align}
As multiple linear transformations in sequence can be collapsed to a single one,
 $\vec f^*(\vec x; \vec\theta)$ can be written as
\begin{align} 
\label{eq:collapse}
    \vec f^*(\vec x; \vec \theta) &= \mat w_{1\rightarrow L}(\vec x)\,\vec x\; \text{,} \\
    \text{with} \quad
        \mat{W}_{{1}\rightarrow {L}}\left(\vec{x}\right) &= \textstyle\prod_{j={1}}^{L} \widetilde{\mat{W}}_j \left(\vec{a}_{j}\right)\; \text{.}
 \end{align}
Thus, $\mat w_{1\rightarrow L}(\vec x)$ {faithfully summarises} the network computations (\cref{eq:bcos_net}) by a single linear transformation (\cref{eq:collapse}).

To explain an activation (\eg, the class logit), we can now either {directly visualise} the corresponding row in $\mat w_{1\rightarrow L}$, see \cref{fig:global_protos,,fig:intermediate_protos}, or the \emph{contributions} according to $\mat w_{1\rightarrow L}$ coming from individual input dimensions. 
We use the resulting spatial {contributions maps} to quantitatively evaluate the explanations. In detail, the input contributions  $\vec s_j^l(\vec x)$ to neuron $j$ in layer $l$ for an input $\vec x$ are given by the
\begin{align}
    \label{eq:contrib}
\textbf{contribution map}\qquad        \vec{s}_{j}^l(\vec x) = \left[\mat W_{1\rightarrow l} (\vec{x})\right]_j^T \odot \vec x\; \text{,}
    \end{align}
with $[\mat w_{1\rightarrow l}]_j$ denoting the $j$th row in matrix $\mat w_{1\rightarrow l}$; as such, the contribution from a single pixel location ${(x, y)}$ is given by $\sum_c[\vec s_j^l(\vec x)]_{(x, y, c)}$ with $c$ the color channels.
}

\subsubsection{Optimising \bcosnets for classification}
\label{subsubsec:optim}

In the following, we discuss why the linear transformations $\mat W_{1\rightarrow L}$ (\cref{eq:collapse}) can be expected to align with relevant input patterns {for models based on \bcos transformations}.

{For this, first note that the output of each neuron---and thus of each layer---is bounded, cf.~\cref{eq:bound,,eq:bcos_suppress}. Since the output of a \bcosnet is computed as a sequence of such bounded transformations, see \cref{eq:linear_sequence}, the output of the network as a whole is also bounded. Secondly, note that a \bcosnet only achieves its upper bound for a given input if every unit achieves its upper bound. Importantly, the individual units can only achieve their maxima by aligning with their inputs (cf.~\cref{eq:bound}). Hence, optimising a \bcosnet to maximise its output over a set of inputs will optimise the model weights to align with those inputs.}

To take advantage of this for \bcos classification models, we train them with the binary cross entropy (BCE) loss
\begin{align} 
    \label{eq:loss}
    \mathcal{L}(\vec x_i, \vec y_i) = \text{BCE}\left(\sigma(\vec f^*(\vec x_i; \vec \theta)/T{ + \vec b}), \vec y_i\right)
\end{align}
for input $\vec x_i$ and the respective one-hot encoding of the label $\vec y_i$; see below (target encoding) for an additional discussion. Here, $\sigma$ denotes the sigmoid function, $\vec b$ and $T$ fixed bias and scaling terms, and $\vec \theta$ the model parameters. 
Note that we choose the BCE loss as it directly entails maximisation. Specifically, in order to reduce the BCE loss, the network is optimised to maximise the (negative) class logit for the correct (incorrect) classes. As discussed in the previous paragraph, this will optimise the weights in each layer of the network to align with their inputs. In particular, they will need to align with class-specific input patterns such that these result in large outputs for the respective class logits.
 
Finally, note that increasing B allows to specifically reduce the output of badly aligned weights in each layer ({\ie, with low cosine similarity to the input}). This decreases the layer's output strength and thus the output of the network as a whole, which increases the alignment pressure during optimisation (thus, higher B$\rightarrow$higher alignment, see \cref{fig:b-abl-quali}).

\myparagraph{Target encoding and logit bias $\vec b$.}
{As discussed, BCE optimises the models to maximise the absolute magnitude of both the positive (target class) as well as the negative logits (other classes). To encourage the models to find \emph{positive} evidence (\ie contributions as defined in \cref{eq:contrib}) for the target class, rather than negative evidence for the other classes, we set the logit bias $\vec b$ to $\log\left(1/(C-1)\right)$ for $C$ classes, which corresponds to a default class probability of 1/C for each class. As a result, the magnitude of $\vec f^\ast (\vec x_i; \vec \theta)$ can be small for non-target classes without incurring a large loss, while it has to be large for the target class to achieve a class confidence close to 1 and thus reduce the BCE loss.

Interestingly, in our experiments we found that models with normalisation layers (see \cref{subsubsec:normed_bcos}) can still exhibit a tendency to focus on negative evidence, especially when the number of classes $C$ is large (\eg on ImageNet), see \cref{subsec:biases}. To overcome this, for our ImageNet experiments we additionally modify the target encoding $\vec y_i$ to be $1/C$ for all non-target classes, which encourages the models to produce weights that are orthogonal to the input for non-target classes, \ie the optimum can only be reached if $[\vec f^\ast(\vec x_i;\vec \theta)]_c\myeq0$ for all $c$ that are not the target class.
}

\subsection{Advanced \bcosnets}
\label{subsec:deep_bcos}
To test the generality of our approach, we investigate how integrating \bcos transformations into conventional DNNs affects their classification performance and interpretability. To do this, in the following we discuss how to combine \bcos layers and their explanations with other commonly used model components, such as activation functions (\cref{subsubsec:activations}), normalisation (\cref{subsubsec:normed_bcos}), and attention layers (\cref{subsubsec:attention}).

Note that whenever converting a linear / convolutional layer in a given network to the corresponding \bcos layer, \textbf{we remove all bias terms} to ensure that the model output is given by a dynamic linear transformation as in \cref{eq:collapse}. For a discussion on the impact of biases, see \cref{subsec:biases}.

{
\subsubsection{Activation functions in \bcos networks}
\label{subsubsec:activations}
Since \bcos transformations themselves are non-linear, \bcos DNNs do not require activation functions to model non-linear relationships (see \cref{subsec:bcos_net}). Hence, most of our models in \cref{sec:results} do not employ any activation functions. Nonetheless, combining \bcos networks with activation functions can still be beneficial. In particular, as we discuss in \cref{subsec:plain_results}, adding an additional non-linearity allows us to study the effect of the parameter B in \cref{eq:bcos} over a wide range of values, including B$\myeq1$, and thus to directly compare them to conventional piece-wise linear models\footnote{For $\text{B}\myeq1$ linear and \bcos transformations are equivalent (\cref{eq:bcos2}).}.

 While there are many potential non-linearities to choose from, in this work, we specifically explore the option of combining the \bcos transformation with MaxOut~\cite{goodfellow2013maxout}. For this, we model every neuron by 2 \bcos transformations %
 of which only the maximum is forwarded to the next layer:
\begin{align} 
    \label{eq:mbcos}
    \text{\mbcos} (\vec x) = \textstyle \max_{i\in\{1, 2\}} \left\{\text{\bcos}(\vec x; \vec w_i)\right\} \;.
\end{align}
We do so for several reasons. First, this operation maintains the models' dynamic linearity and is thus compatible with the \bcos explanations. Secondly, as discussed above, for B$\myeq1$ the resulting model is a piece-wise linear model, which allows us to interpolate between conventional piece-wise linear models and the proposed \bcos models. %
Lastly, while the ReLU~\cite{nair2010relu} operation also fulfills these properties, we noticed that \bcos models without any normalisation layers were much easier to optimise with MaxOut instead of ReLU, which we attribute to avoiding `dying neurons' \cite{goodfellow2013maxout}. 
}

{
\subsubsection{Adapting normalisation layers for \bcosnets}
\label{subsubsec:normed_bcos}
In this section we discuss how normalisation layers can be integrated into \bcosnets. Specifically,  we note that these layers (\eg BatchNorm~\cite{ioffe2015batchnorm}, LayerNorm~\cite{ba2016layer}, PositionNorm~\cite{li2019positional}, or InstanceNorm~\cite{ulyanov2016instance}) are (1) non-linear operations and (2) can pose a challenge to the \emph{completeness} of the explanations. By interpreting them as dynamic linear functions and removing bias terms, we can nonetheless seamlessly integrate them into \bcos explanations.
First, however, let us define the normalisation layers.

\myparagraph{Definition.} Normalisation layers have been shown to benefit neural network training by whitening the input $\vec x$ via
\begin{align}
    \label{eq:norm_def}
    \ast\text{Norm} (\vec x, \inputset{X};\gamma, \beta) = \frac{\vec x - \langle\inputset{X}\rangle_\ast}{\sqrt{\text{var}_\ast(\inputset{X})}} \times \vec \gamma + \vec \beta\;,
\end{align}
with $\inputset{X}\myin \mathbb R^{b\times c\times h\times w}$ a batch of $b$ inputs, each with $c$ channels and a width $w$ and a height $h$; further, $\vec x\myin \mathbb R^{c}$ is a single representation vector from a specific input at a given location.
As indicated by the placeholder $\ast$, the individual normalisation layers differ mainly in the choice of dimensions over which the mean $\langle \cdot \rangle$ and the variance $\text{var}$ are computed. Finally, note that at inference time, the mean and variance terms are sometimes replaced by running estimates of those values, most commonly done so in BatchNorm~\cite{ioffe2015batchnorm}.

To avoid ambiguities\footnote{Note that LayerNorm has in fact been interpreted differently by different authors (\eg \cite{liu2022convnet} vs.~\cite{li2019positional,wu2018group}), \ie either as \cref{eq:layernorm} or \cref{eq:posnorm}.} and facilitate the comparison of different normalisation schemes, we define the normalisation layers by the respective choice of dimensions they use:
\begin{alignat}{2}
\text{BatchNorm}(\vec x, \inputset{X};\gamma, \beta)& \coloneqq \text{B}&&\text{HW-}\text{Norm} (\vec x, \inputset{X};\gamma, \beta)
\label{eq:batchnorm}
\\
\text{LayerNorm}(\vec x, \inputset{X};\gamma, \beta)& \coloneqq \text{C}&&\text{HW-}\text{Norm} (\vec x, \inputset{X};\gamma, \beta)
\label{eq:layernorm}
\\
\text{InstanceNorm}(\vec x, \inputset{X};\gamma, \beta)& \coloneqq &&\text{HW-}\text{Norm} (\vec x, \inputset{X};\gamma, \beta)
\label{eq:instancenorm}
\\
\text{PositionNorm}(\vec x, \inputset{X};\gamma, \beta)& \coloneqq \text{C}&&\text{\phantom{HW}-}\text{Norm} (\vec x, \inputset{X};\gamma, \beta)
\label{eq:posnorm}
\end{alignat}
\ie BatchNorm estimates mean and variance for each channel independently (no 'C') over all spatial dimensions (HW) and inputs in the batch (B), whereas LayerNorm computes mean and variance across all activations (CHW) in a single input, but does not use other inputs to estimate those values.%

By defining the normalisation layers as above, an additional candidate naturally emerges, which we call \emph{AllNorm}:
\begin{alignat}{2}
\label{eq:allnorm}
\text{AllNorm}(\vec x, \inputset{X};\gamma, \beta)& \coloneqq \text{BC}&&\text{HW-}\text{Norm} (\vec x, \inputset{X};\gamma, \beta)\,.
\end{alignat}

\myparagraph{Dynamic linear normalisation.} Normalising the input by the variance of course constitutes a highly non-linear operation. To nonetheless integrate this operation into the \bcos-based dynamic linear explanations (cf.~\cref{eq:linear_sequence}), we interpret the normalisation itself to be a dynamic linear function with:
\begin{align}
    \vec w_{\ast\text{Norm}} (\vec x, \inputset{X}) &= \gamma \times {\sqrt{\text{var}^{-1}_\ast(\inputset{X})}}\\
    \text{s.t.}\; \ast\text{Norm} (\vec x, \inputset{X};\gamma, \beta) &= \vec w_{\ast\text{Norm}} \mytimes\left( {\vec x - \langle\inputset{X}\rangle_\ast}\right) + \vec \beta\;,
\end{align}

\myparagraph{Explanation completeness.} As becomes clear from \cref{eq:norm_def}, normalisation layers introduce additive biases into the model prediction (\ie $\beta$ and, if running estimates are used, by the running mean estimation of $\langle \inputset{X}\rangle_\ast$). The model output is thus not given by a dynamic linear transformation anymore (cf.~\cref{eq:collapse}), but instead by an \emph{affine} transformation $\vec f^\ast (\vec x;\vec \theta)\myeq \mat W(\vec x) \vec x + \vec b(\vec x)$ with an input-dependent bias $\vec b(\vec x)$.

The contribution maps in \cref{eq:contrib} would thus not be \emph{complete} {(\ie the sum over contributions would not yield the respective class logit)}, as they do not take the biases into account. This, however, is undesirable for model-faithful explanations \cite{srinivas2019full,sundararajan2017axiomatic}, as the bias terms can play a significant role in the model decision (see \cref{subsec:biases}).

To alleviate this, we use bias-free (BF) normalisation~\cite{mohan2019robust}, and remove the \emph{learnt} biases $\beta$, {which would make a linear summary of the model via $\mat W(\vec x)\vec x$ incomplete}\footnote{Note that the {subtraction of the mean $\langle \inputset{X} \rangle_*$ }does not introduce \emph{external} biases that would be unaccounted for in the explanations, as this operation can be modelled by a linear transformation of the input.}:
\begin{align}
    \label{eq:norm_def_ours}
    \ast\text{Norm}^\text{BF} (\vec x, \inputset{X};\gamma, \beta) = \frac{\vec x - \langle\inputset{X}\rangle_\ast%
    }{\sqrt{\text{var}_\ast(\inputset{X})}} \times \vec \gamma + {\color{red} 0}\;.
\end{align}
Moreover, for normalization layers that replace the mean computation $\langle\inputset{X}\rangle_\ast$ by a running mean estimate at inference time (\ie BatchNorm), we additionally remove the centering term $\langle\inputset{X}\rangle_\ast$, since otherwise external biases would be introduced at inference time.

\subsubsection{\bcos ViTs: combining \bcos and attention layers}
\label{subsubsec:attention}
Thus far, our discussion focused on designing interpretable convolutional neural networks (CNNs), which have long dominated computer vision research. Recently, however, CNNs are often surpassed by transformers~\cite{vaswani2017attention}, which---if the current development is any indication---will replace CNNs for ever more tasks and domains. In the following, we therefore investigate how to design \bcos transformers.

For this, first note that the core difference between vision transformers (ViTs) and CNNs is the use of attention layers and additive positional encodings. The remaining ViT components (tokenisation, MLP blocks, normalisation layers, classification head) have a direct convolutional counterpart. 

In particular, the tokenisation module is typically given by a $K\mytimes K$ convolution with a stride and kernel size of $K$, the MLP blocks effectively perform 1x1 convolutions (cf.~\cite{liu2022convnet}), the normalisation layers can be expressed as PositionNorm in CNNs (\cref{eq:posnorm}) and the classification head is either given by a single or multiple linear layers. 

Therefore, in the following, we discuss the attention mechanism and the positional embeddings in more detail.

\myparagraph{Attention.} Interestingly, the attention operation itself is already dynamic linear and attention thus lends itself well to be integrated into the linear summary according to \cref{eq:collapse}:
\begin{align}
\label{eq:attention}
    \text{Att}({{\mat X}}; \mat Q,\mat K,\mat V) 
    \;&=\;
    \underbrace{\text{softmax}\left(\mat X^T\mat q^T {{\mat K}} {{\mat X}}\right)}_{\text{Attention matrix $\mat A({{\mat X}})$}}\underbrace{\,\mat v{{\mat X}}\,
    \vphantom{\text{softmax}\left(\mat q {{\mat X}} 
    {{\mat X}}^T\mat k^T\right)
    }
    }_{\text{Value}({\mat X})} \\
    \;&=\;
    \underbrace{\mat A({{\mat X}})\, \mat V
    \vphantom{\text{softmax}\left(\mat q {{\mat X}^T} 
    {{\mat X}}\mat k^T\right)
    }
    }_{\mat w({{\mat X}})}\, {{\mat X}}\; 
    \;=\; \mat W({{\mat X}}){{\mat X}}.
\end{align}
Here, $\mat q,\mat k$ and $\mat V$ are the query, key, and value transformation matrices and $\mat X$ denotes the matrix of input tokens to the attention layer. Softmax is computed column-wise.

In multi-head self-attention (MSA), $H$ attention heads are used in parallel. Their concatenated outputs are then linearly projected by a matrix $\mat U$ via
\begin{align}
    \text{MSA}({\mat X}) = \mat U \left[\mat W_1 ({\mat X}) {\mat X},\; ...\;, \mat W_H ({\mat X}) {\mat X}\right]\,,
\end{align}
which is still dynamic linear. While there are multiple linear operations involved in MSA (query, key, value, and projection computations) and therefore various options for introducing \bcos layers, we empirically found not replacing\footnote{I.e., these transformations effectively use a \bcos layer with B$\myeq1$.} the query, key, and value computations to yield the best results, which we attribute to a potentially higher compatibility with the softmax function. Hence, to adapt the attention layers to the \bcos framework, we only replace the linear projection via $\mat U$ by a \bcos transformation in our experiments.

\myparagraph{Positional encoding.} In contrast to CNNs, which possess a strong inductive bias regarding spatial relations (local connectivity), transformers are invariant with respect to the token order and thus lack such a `locality bias'. To nevertheless leverage spatial information, it is common practice to break the symmetry between tokens by adding a (learnt) embedding $\mat E$ to the input tokens $\mat X$ such that $\mat x'\myeq \mat x + \mat e$.

While explanations solely with respect to $\mat X$ are not thus complete (cf.~\cref{subsubsec:normed_bcos}), since part of the model output will be based on contributions from $\mat E$, we experimentally find that the positional embeddings do not negatively impact the interpretability of the ViTs. How to design ViT models without the need for such external biases represents an interesting direction for future work.
}

\section{Experimental setting}
\label{sec:experiments}

\myparagraph[0]{Datasets.}
We evaluate the accuracies of a wide range of \bcos networks on the CIFAR-10~\cite{krizhevsky2009cifar10} and the ImageNet~\cite{deng2009imagenet} datasets. We use the same datasets for the qualitative and quantitative evaluations of the model-inherent explanations.

\myparagraph{CIFAR10 models.}
For the CIFAR10 experiments, we develop a simple fully-convolutional \bcos DNN, consisting of 9 convolutions, each with a kernel size of 3, followed by a global pooling operation. We evaluate a network without additional non-linearities as well as with MaxOut units, see \cref{subsubsec:activations}. 
{Additionally, to study the normalisation layers and the effect of the bias terms (cf.~\cref{eq:norm_def_ours}), we evaluate various \bcos ResNets. For this we adapt a conventional ResNet-56 \cite{he2016deep} by removing all activation functions and bias parameters and replacing all linear / convolutional layers by their corresponding \bcos version. Additionally, we replace the conventional batch normalisation layers by the (modified) normalisation layers as described in \cref{subsubsec:normed_bcos}.}

{\myparagraph{ImageNet models.}
For the ImageNet experiments, we rely on the publicly available~\cite{pytorch} implementations of a wide range of CNNs (VGGs~\cite{simonyan2015vgg}, ResNets~\cite{he2016deep}, DenseNets~\cite{huang2017densely}, ResNext~\cite{xie2017aggregated} and ConvNext~models~\cite{liu2022convnet}). Further, we evaluate \bcos versions of the Vision Transformers (ViTs) \cite{dosovitskiy2021an}, specifically the version from~\cite{beyer2022better}, as well as ViT$_C$ models as in \cite{xiao2021early}, \ie ViTs with a shallow stem of four convolutional layers. We adapt all of those architectures to \bcos networks as described in \cref{subsec:deep_bcos} and train them according to standard protocols. Specifically, we train most CNNs with the default 90 epochs training paradigm as found in the torchvision library~\cite{torchvision}; additionally, we evaluate various models when employing a more sophisticated training recipe \cite{torchvisionConvnext} of 600 epochs, based on the ConvNext~\cite{liu2022convnet} training protocol. Finally, for the ViT models, we follow the protocol proposed by \cite{beyer2022better}, which has shown strong performance for ViTs with only 90 epochs of training.  For more details on the training procedure, see supplement (Sec.~C).}

\myparagraph{Image encoding.} We add three additional channels and encode images as \mbox{[$r$, $g$, $b$, $1\myminus r$, $1\myminus g$, $1\myminus b$]}, with $r, g, b\,\myin[0, 1]$ the red, green, and blue color channels. On the one hand, this reduces a bias towards bright regions in the image\footnote{
{Note that the model output is given by a weighted sum over the input (\cref{eq:collapse}).}
Since black pixels are encoded as zero in the conventional encoding, they cannot contribute to the class logits (cf.~\cref{eq:contrib}).} \cite{Boehle2021CVPR}. On the other hand, colors with the \emph{same angle in the original encoding}---\ie, $[r_1, g_1, b_1]\mypropto[r_2, g_2, b_2]$---\emph{are unambiguously encoded by their angles under the new encoding}. Therefore, the linear transformation $\mat w_{1\rightarrow l}$ can be decoded into colors just based on the angles of each pixel, see \cref{fig:global_protos}. For a detailed discussion, see supplement (Sec.~D).

\begin{figure}
    \centering
    \begin{subfigure}[b]{\linewidth}
    \includegraphics[width=\linewidth, trim=1em 1em .5em 0, clip]{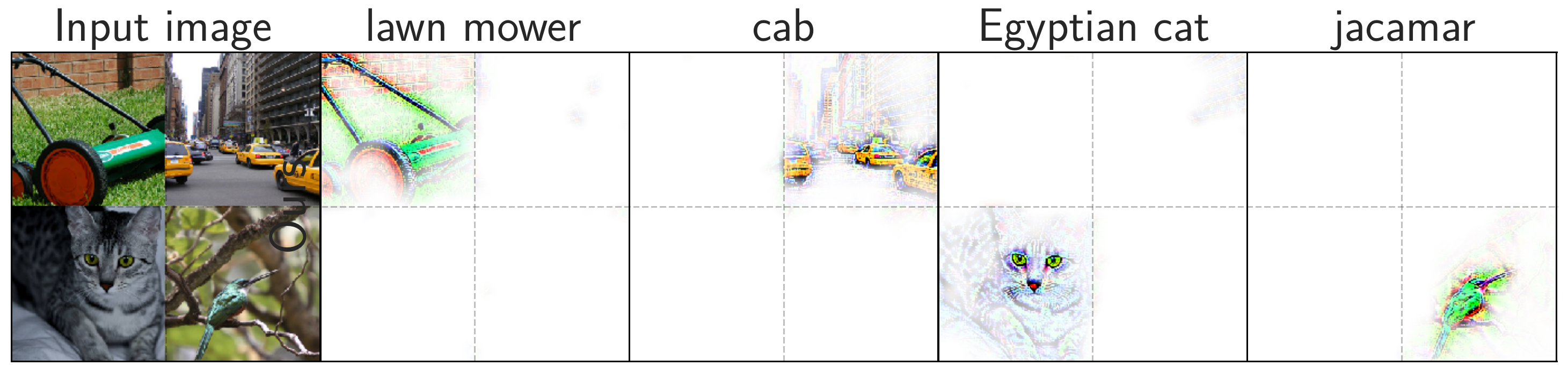}
    \end{subfigure}
    \caption{2$\times$2 pointing game example. \textbf{Column 1}: input image. \textbf{Columns 2--5}: explanations for individual class logits.}
    \label{fig:local_quali}
\end{figure}

\myparagraph{Evaluating explanations.} To compare explanations for the model decisions and {evaluate their faithfulness}, we employ the \emph{grid pointing game} \cite{Boehle2021CVPR}. That means we evaluate the trained models on a synthetic 3x3 grid of images of different classes and for each of the corresponding class logits measure how much positive attribution an explanation method assigns to the correct location in the grid, see \cref{fig:local_quali} for an example. {Note that due to the global attention in ViTs such grids are much further out of distribution at every layer of the model than they are for CNNs, for which the locally applied convolutional kernels are not affected by the synthetic nature of the input grids. The grid pointing game, however, relies on the assumption that the model extracts similar features in the synthetic setting and \emph{locally} correctly classifies the subimages. To ensure that this assumption (at least approximately) holds, we therefore apply the ViTs in a sliding window fashion to the synthetic image grids, \ie we apply the ViTs to patches of size 224x224 extracted at a stride of 112 from the image grid. As this significantly increases the computational cost, we use 2x2 grids for ViTs.}

Following \cite{Boehle2021CVPR}, we construct 500 grids from the most confidently and correctly classified images and compare the {model-inherent} contribution maps (\cref{eq:contrib}) against several commonly employed {post-hoc} explanation methods under two settings. First, we evaluate all methods on \bcos networks to investigate which method provides the best explanation \emph{for the same model}. Secondly, we further evaluate the post-hoc methods on pre-trained versions of the original models. This allows to compare explanations \emph{between different models} and assess the \emph{explainability gain} obtained by converting conventional models to \bcos networks. Lastly, all attribution maps (except for \gcam, which is of much lower resolution to begin with) are smoothed by a 15$\times$15 (3$\times$3) kernel to better account for negative attributions in the localisation metric for ImageNet (CIFAR-10) images, which is small with respect to the image size of 224$\mytimes$224 (32$\mytimes$32).

\myparagraph{Visualisations details.} 
{For generating the visualisations of the linear transforms for individual neurons $n$ in layer $l$ (cf.~\cref{fig:global_protos,,fig:intermediate_protos}), we proceed as follows. First, we select all pixel locations  $(x, y)$ that positively contribute to the respective activation (\eg, class logit) as computed by \cref{eq:contrib};
\ie, $\{(x, y)\; \text{s.t.}\, \sum_c\, [\vec s_n^l(\vec x)]_{(x, y, c)}\mygreater0\}$ with $c$ the 6 color channels (see image encoding). Then, we normalise the weights of each color channel such that the corresponding weights (\eg, for $r$ and $1\myminus r$) sum to 1.
Note that this normalisation maintains the angle for each color channel pair (\ie, $r$ and $1\myminus r$), but produces values in the allowed range $r,g,b\,$%
$\in\,$%
$[0,1]$. These normalised weights can then directly be visualised as color images. The opacity of a pixel is set to $\min(\lVert\vec w_{(x, y)}\rVert_2/p_{99.9}, 1)$, with $p_{99.9}$ the $99.9$th percentile over the weight norms $\lVert\vec w_{(x, y)}\rVert_2$ across all $(x, y)$.
}
\section{Results}
\label{sec:results}

In this section, we analyse the performance and interpretability of \bcos models. For this, in \cref{subsec:plain_results} we show results of `simple' \bcos models without advanced architectural elements such as skip connections. In this context, we investigate how the B parameter influences \bcos models in terms of performance and interpretability.
Thereafter, in \cref{subsec:advanced_results}, we present quantitative results of the \emph{advanced} \bcos models, i.e., \bcos models based on common DNN architectures (cf.~\cref{subsec:deep_bcos}). In \cref{subsec:quali_results}, we visualise and qualitatively discuss explanations for individual neurons of the advanced \bcos models, and finally, in \cref{subsec:biases}, we discuss the impact of the bias terms in the normalisation layers.

\subsection{Simple \bcos models}
\label{subsec:plain_results}
In the following, we discuss the experimental results of simple \bcos models evaluated on the CIFAR-10 dataset. 
\begin{table}[h]
    \centering
    {\setlength{\tabcolsep}{0.3em}\setlength\extrarowheight{2pt}
    \begin{tabular}{>{\centering\arraybackslash}  
    >{\centering\arraybackslash}p{1.7cm} |
    >{\centering\arraybackslash}p{.6cm} 
    >{\centering\arraybackslash}p{.6cm} 
    >{\centering\arraybackslash}p{.6cm} 
    >{\centering\arraybackslash}p{.6cm} 
    >{\centering\arraybackslash}p{.6cm} 
    >{\centering\arraybackslash}p{.6cm} 
    >{\centering\arraybackslash}p{.6cm} |
    >{\centering\arraybackslash}p{1cm} 
    }
   \multicolumn{1}{c}{} &\multicolumn{7}{c}{\footnotesize \bf MaxOut \bcos networks}&{\footnotesize \bf plain}
    \\
    \footnotesize \textbf{B}
    & {\footnotesize 1.00}
    & {\footnotesize 1.25}
    & {\footnotesize 1.50}
    & {\footnotesize 1.75}
    & {\footnotesize 2.00}
    & {\footnotesize 2.25}
    & {\footnotesize 2.50}
    & {\footnotesize 2.00}
    \\\hline
    \footnotesize \textbf{Accuracy (\%)}
    &\footnotesize 93.5 %
    &\footnotesize 93.8 %
    &\footnotesize 93.7 %
    &\footnotesize 93.7 %
    &\footnotesize 93.2 %
    &\footnotesize 92.6 %
    &\footnotesize 92.4 %
    &{\footnotesize 91.5} %
    \end{tabular}}
    \caption{\textbf{CIFAR-10} accuracy of a \bcos model
    without additional non-linearities (\textbf{plain}) and for \bcos models with MaxOut (\cref{eq:mbcos}) and increasing values for B (left to right).
    }
    \vspace{-.25em}
    \label{tbl:c10_table}
\end{table}
\begin{figure}
    \centering
    \begin{subfigure}[b]{\linewidth}\centering
    \includegraphics[width=\linewidth]{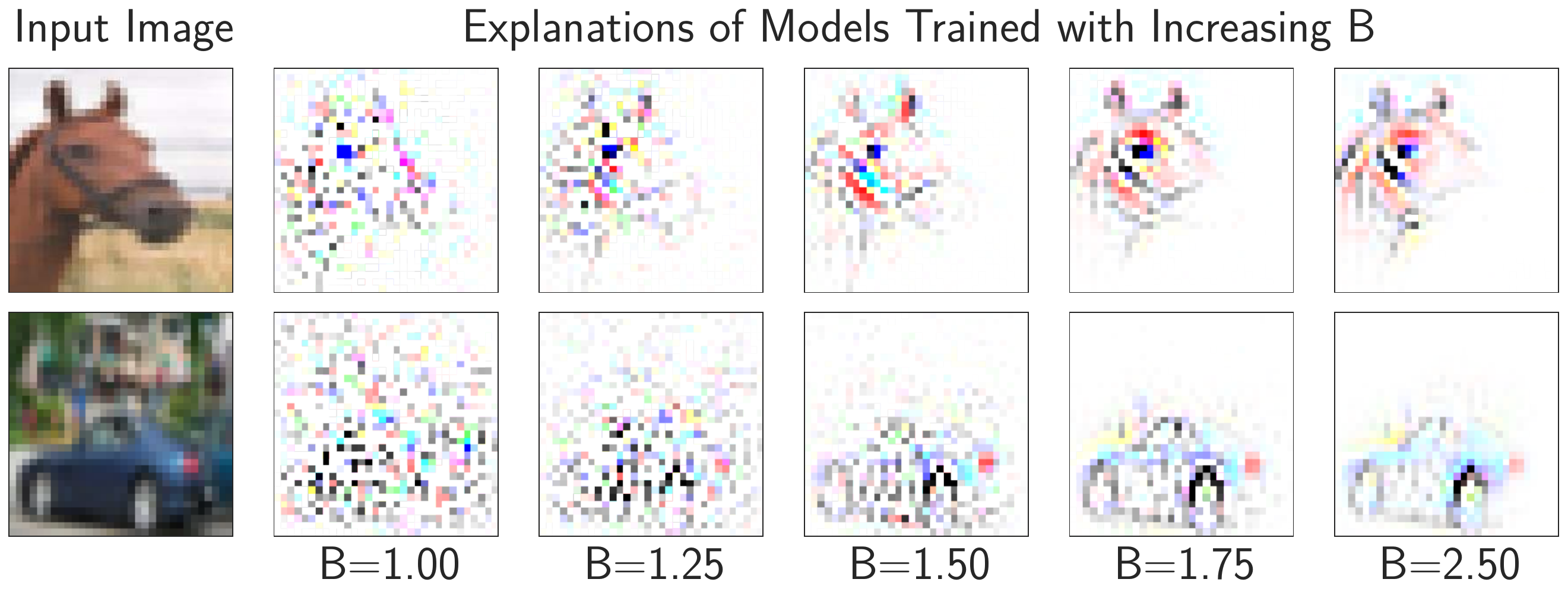}
    \end{subfigure}
    \caption{
    \textbf{Col.~1}: Input images.
    \textbf{Cols.~2-6}: Explanations for different classes $c$ (top: `horse'; bottom: `car') of models trained with increasing B. 
    For higher B, the model-inherent linear explanations $[\mat w_{1\rightarrow l}]_c$ increasingly align with discriminative input patterns, thus becoming more interpretable.
    }
    \label{fig:b-abl-quali}
\end{figure}
\begin{figure}[t]
    \centering
    \begin{subfigure}[b]{1.0\linewidth}
    \includegraphics[width=\textwidth, trim=.5em 1em .5em 2em, clip]{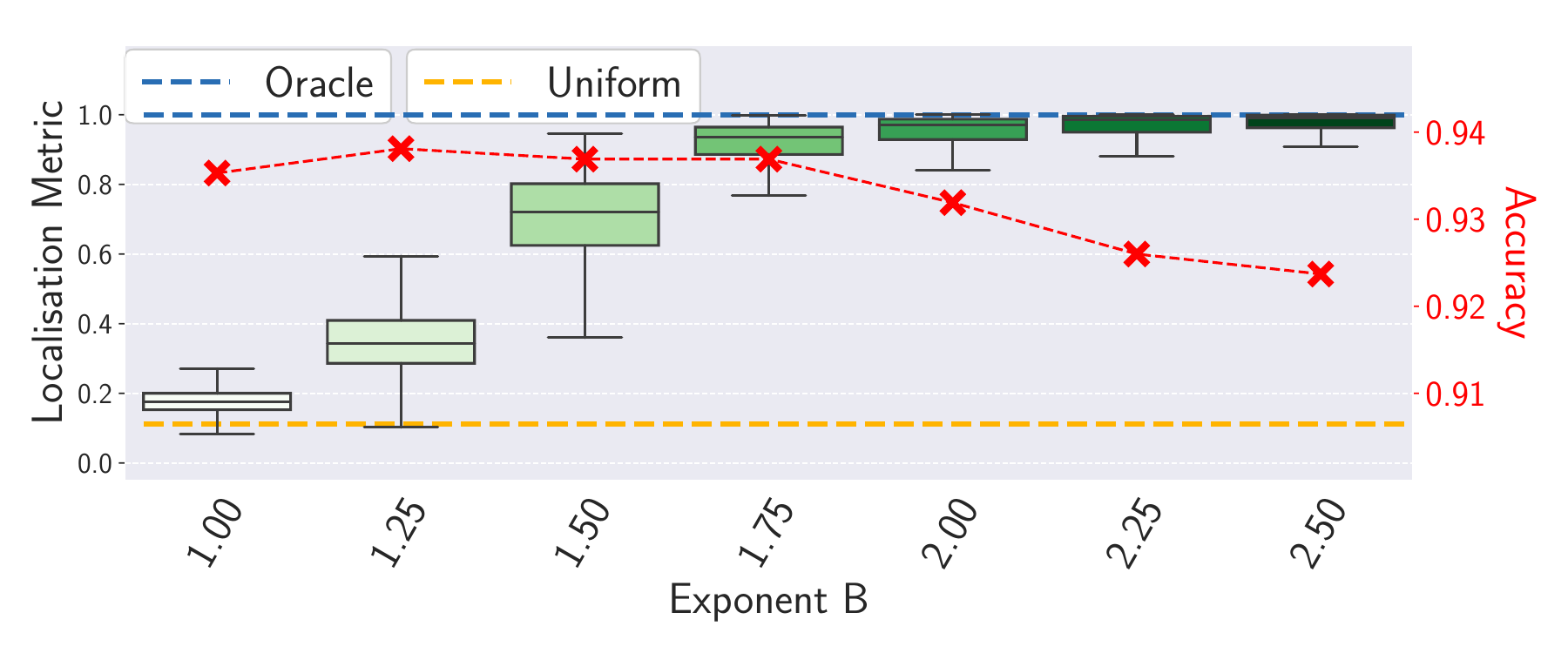}
    \end{subfigure}
    \caption{Accuracy (crosses) and localisation (box plots) results for a \bcos network trained with different B. While decreasing accuracy, larger B significantly improve localisation.}
    \label{fig:local_c10}
\end{figure}

\myparagraph{Accuracy.}
In \cref{tbl:c10_table}, we present the test accuracies of various \bcos models trained on CIFAR-10. We show that a {plain \bcos model} ($\text{B}\myeq2$) without any add-ons (ReLU, batch norm, etc.) can achieve competitive\footnote{A ResNet-20 achieves 91.2\%~\cite{he2016deep} with the same data augmentation.} performance. 
By modelling each neuron via 2 MaxOut units (\cref{eq:mbcos}), the performance can be increased and the resulting model ($\text{B}\myeq2$) performs on par with a ResNet-56 (achieving 93.0\%, see \cite{he2016deep}). Further, we see that an increase in the parameter B leads to a decline in performance from $93.8\%$ for 
$\text{B}\myeq1.25$ to $92.4\%$  for 
$\text{B}\myeq2.5$. Notably, despite its simple design, our strongest model with $\text{B}\myeq1.25$ performs similarly to the strongest ResNet model ($93.6\%$) reported in \cite{he2016deep}.

\myparagraph{Model interpretability.}
As discussed in \cref{subsubsec:optim}, we expect an increase in B to increase the alignment pressure on the weights during optimisation and thus influence the models' optima, similar to the single unit case in \cref{fig:toy_example}. 
This is indeed what we observe. For example, in \cref{fig:b-abl-quali}, we visualise $[\mat W_{1\rightarrow l}(\vec x_i)]_{y_i}$ (see \cref{eq:collapse}) for different samples $i$ from the CIFAR-10 test set. For higher values of B, the weight alignment increases notably from piece-wise linear models (B=1) to \bcos models with higher B (B=2.5). Importantly,  this does not only lead to an increase in the {visual quality} of the explanations, but also to quantifiable gains in model interpretability. In particular, as we show in \cref{fig:local_c10}, the spatial contribution maps defined by $\mat W_{1\rightarrow l}(\vec x_i)$ (see \cref{eq:contrib}) of models with larger B values score significantly higher in the localisation metric (see \cref{sec:experiments}).

\subsection{Advanced \bcos models}
\label{subsec:advanced_results}
\myparagraph{\bcos CNNs -- Accuracy.} In \cref{tbl:cnn:90epochs,,tbl:cnn:600epochs}, we present the top-1 accuracies of the \bcos models on the ImageNet validation set for the 90 epoch and 600 epoch training paradigms respectively. 
For comparison, we also show the difference to the accuracy of the respective baseline models as reported in \cite{torchvision} ($\Delta^1$). As can be observed, the \bcos models are highly competitive, achieving accuracies on par with the baselines for some models (ResNet-50, $\Delta^1\myeq\myminus0.2$), and a worst-case drop of $\Delta^1\myeq\myminus2.0$ (ResNext-50-32x4d) under the 90 epoch training paradigm (\cref{tbl:cnn:90epochs}). Interestingly, we find that the differences to the baseline models vanish when training the baselines without bias terms in the convolution and normalisation layers, \ie as is done in \bcos models; we denote the difference to the baselines without biases by $\Delta^2$. 

When employing a long training protocol with additional data augmentation as in \cite{liu2022convnet}, the performance of the \bcos models can be further improved: \eg, the \bcos ResNet-152 increases its top-1 accuracy by $3.2$pp (compare \cref{tbl:cnn:90epochs,,tbl:cnn:600epochs}) and achieves a top-1 accuracy of $80.2\%$, which is an unprecedented performance for models that inherently provide such detailed explanations (see \cref{subsec:quali_results}). Nonetheless, especially for the ConvNext models, we observe a significant performance drop with respect to the baseline numbers reported in \cite{torchvision} (up to $5.0$pp for ConvNext-Tiny). Note, however, that the training protocol has been meticulously optimised for ConvNext models in \cite{liu2022convnet}; as such, we expect that the performance of \bcos models will significantly improve if a similar effort as in \cite{liu2022convnet} is undertaken for optimising the model architecture and training procedure.
\begin{table}[h!]
\centering
\addtolength{\tabcolsep}{-1pt} 
\begin{tabular}{lccc|ccc}
 & \multicolumn{3}{c|}{\textbf{Accuracy}} & \multicolumn{3}{c}{\textbf{Localisation}} \\[.5em]
\textbf{\bcos Model}& \% & $\Delta^1$ & $\;\Delta^2$ & \%$\,$ & $\;\Delta_\text{IntG}^\text{\bcos}$ & $\Delta_\text{IntG}^\text{basel.}$ \\
\midrule
VGG-11 &{\scriptsize  69.0} &{\scriptsize \color{red} -1.4} &{\scriptsize \color{red} -0.6} &{\scriptsize  85.6} &{\scriptsize \color{teal} +26.4} &{\scriptsize \color{teal} +64.5} \\
ResNet-18 &{\scriptsize  68.7} &{\scriptsize \color{red} -1.1} &{\scriptsize \color{teal} +0.3} &{\scriptsize  86.9} &{\scriptsize \color{teal} +34.0} &{\scriptsize \color{teal} +64.0} \\
ResNet-34 &{\scriptsize  72.1} &{\scriptsize \color{red} -1.2} &{\scriptsize \color{red} -0.1} &{\scriptsize  89.0} &{\scriptsize \color{teal} +36.9} &{\scriptsize \color{teal} +65.2} \\
ResNet-50 &{\scriptsize  75.9} &{\scriptsize \color{red} -0.2} &{\scriptsize \color{teal} +2.4} &{\scriptsize  90.2} &{\scriptsize \color{teal} +32.7} &{\scriptsize \color{teal} +63.3} \\
ResNet-101 &{\scriptsize  76.3} &{\scriptsize \color{red} -1.1} &{\scriptsize \color{teal} +6.2} &{\scriptsize  91.8} &{\scriptsize \color{teal} +33.9} &{\scriptsize \color{teal} +64.4} \\
ResNet-152 &{\scriptsize  76.6} &{\scriptsize \color{red} -1.7} &{\scriptsize \color{teal} +3.2} &{\scriptsize  91.3} &{\scriptsize \color{teal} +33.6} &{\scriptsize \color{teal} +63.9} \\
DenseNet-121 &{\scriptsize  73.6} &{\scriptsize \color{red} -0.8} &{\scriptsize \color{teal} +0.2} &{\scriptsize  92.1} &{\scriptsize \color{teal} +41.2} &{\scriptsize \color{teal} +69.1} \\
DenseNet-161 &{\scriptsize  76.6} &{\scriptsize \color{red} -0.5} &{\scriptsize \color{teal} +5.2} &{\scriptsize  93.4} &{\scriptsize \color{teal} +47.1} &{\scriptsize \color{teal} +68.5} \\
DenseNet-169 &{\scriptsize  75.0} &{\scriptsize \color{red} -0.6} &{\scriptsize \color{teal} +0.3} &{\scriptsize  91.8} &{\scriptsize \color{teal} +49.3} &{\scriptsize \color{teal} +67.7} \\
DenseNet-201 &{\scriptsize  75.6} &{\scriptsize \color{red} -1.3} &{\scriptsize \color{teal} +0.3} &{\scriptsize  93.0} &{\scriptsize \color{teal} +44.8} &{\scriptsize \color{teal} +68.8} \\
ResNext-50-32x4d &{\scriptsize  75.6} &{\scriptsize \color{red} -2.0} &{\scriptsize \color{teal} +0.6} &{\scriptsize  91.2} &{\scriptsize \color{teal} +28.2} &{\scriptsize \color{teal} +64.1} \\
\end{tabular}
\addtolength{\tabcolsep}{+1pt} 
\caption{%
{
\textbf{Left:} Top-1 accuracies on ImageNet for a standard 90 epoch training protocol \cite{torchvision}. $\Delta^1$: Accuracy difference with respect to the numbers reported in \cite{torchvision}. $\Delta^2$: Accuracy difference with respect to baselines trained without biases. This setting puts the baselines and the \bcos models on an equal footing, as \bcos models do not use biases either. \textbf{Right:} Additionally, we show localisation scores of the model-inherent explanations (\cref{eq:contrib}) as well as localisation scores of IntGrad evaluated on the \bcos models ($\Delta^\text{\scriptsize \bcos}_\text{\scriptsize IntG}$) and the pre-trained baselines ($\Delta^\text{\scriptsize basel.}_\text{\scriptsize IntG}$) as obtained from \cite{torchvision}. We observe significant localisation gains for all models.
}}
\label{tbl:cnn:90epochs}
\end{table}
\begin{table}[h!]
\centering
\begin{tabular}{lcc|ccc}
 & \multicolumn{2}{c|}{\textbf{Accuracy}} & \multicolumn{3}{c}{\textbf{Localisation}} \\[.5em]
\textbf{\bcos Model}& \% & $\Delta^1$ & \%$\,$ & $\;\Delta_\text{IntG}^\text{\bcos}$ & $\Delta_\text{IntG}^\text{basel.}$ \\
\midrule
ResNet-50 &{\scriptsize  79.5} &{\scriptsize \color{red} -1.4} &{\scriptsize  86.2} &{\scriptsize \color{teal} +41.0} &{\scriptsize \color{teal} +58.6} \\
ResNet-152 &{\scriptsize  80.2} &{\scriptsize \color{red} -2.1} &{\scriptsize  85.3} &{\scriptsize \color{teal} +44.2} &{\scriptsize \color{teal} +55.5} \\
ConvNext-Tiny &{\scriptsize  77.5} &{\scriptsize \color{red} -5.0} &{\scriptsize  69.6} &{\scriptsize \color{teal} +37.7} &{\scriptsize \color{teal} +46.5} \\
ConvNext-Base &{\scriptsize  79.7} &{\scriptsize \color{red} -4.4} &{\scriptsize  81.4} &{\scriptsize \color{teal} +35.2} &{\scriptsize \color{teal} +54.6} \\
\end{tabular}
\caption{%
{
\textbf{Left:} ImageNet accuracies for the Conv\-Next-inspired training protocol \cite{torchvisionConvnext}. $\Delta^1$: Accuracy difference with respect to the numbers reported in \cite{torchvisionConvnext}. While we observe a significant drop with respect to the ConvNext models, we note that the training protocol has been optimised explicitly for those architectures \cite{liu2022convnet} and that \bcos models do not employ biases (cf.~\cref{tbl:cnn:90epochs}) to ensure $\vec y (\vec x) \myeq \mat w(\vec x)\vec x$. \textbf{Right:} Additionally, we show the localisation scores of the model-inherent explanations (\cref{eq:contrib}) as well as the localisation scores of IntGrad evaluated on the \bcos models ($\Delta^\text{\scriptsize\bcos}_\text{\scriptsize IntG}$) and the pre-trained baselines ($\Delta^\text{\scriptsize basel.}_\text{\scriptsize IntG}$) as obtained from \cite{torchvision}. We observe significant localisation gains for all models.
}}
\label{tbl:cnn:600epochs}
\end{table}

\myparagraph{\bcos CNNs -- Interpretability.} Apart from the accuracy, in \cref{tbl:cnn:90epochs,,tbl:cnn:600epochs} we further provide the mean localisation scores obtained in the grid pointing game (cf.~\cref{fig:local_quali} and \cref{sec:experiments}) of the model-inherent explanations according to \cref{eq:contrib}. Additionally, we show the difference to one of the most popular post-hoc attribution methods (IntGrad) evaluated on the \bcos models ($\Delta^\text{\bcos}_\text{\scriptsize IntG}$) and the baselines ($\Delta^\text{\scriptsize basel.}_\text{\scriptsize IntG}$).

Across all models, we find that the model-inherent explanations achieve significantly higher localisation scores than IntGrad; this difference is even more pronounced when comparing to the IntGrad explanations for the baselines. 

\begin{figure}
    \centering
    \begin{subfigure}[b]{\linewidth}
    \includegraphics[width=\textwidth]{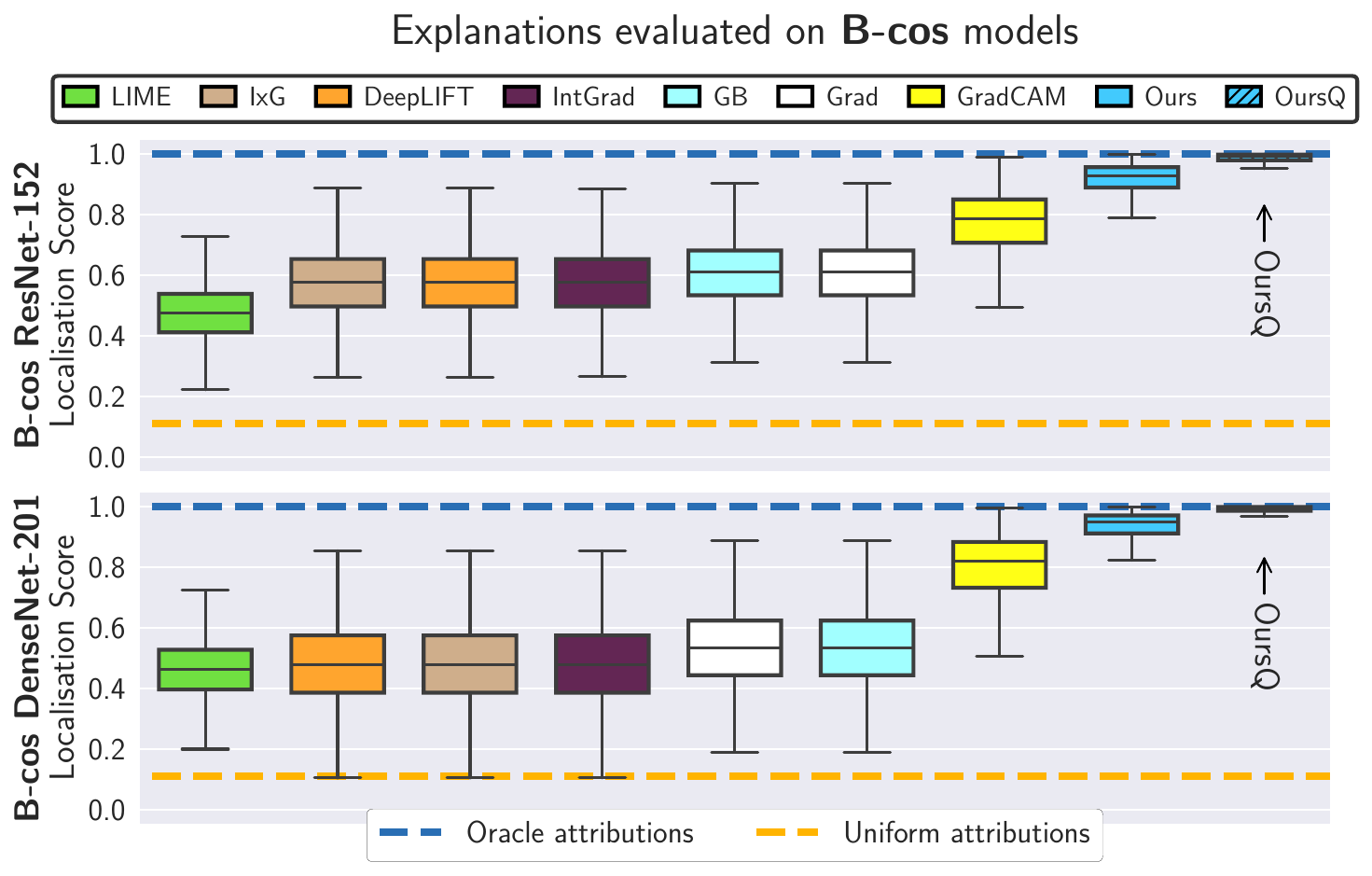}
    \end{subfigure}
    \begin{subfigure}[b]{\linewidth}
    \includegraphics[width=\textwidth]{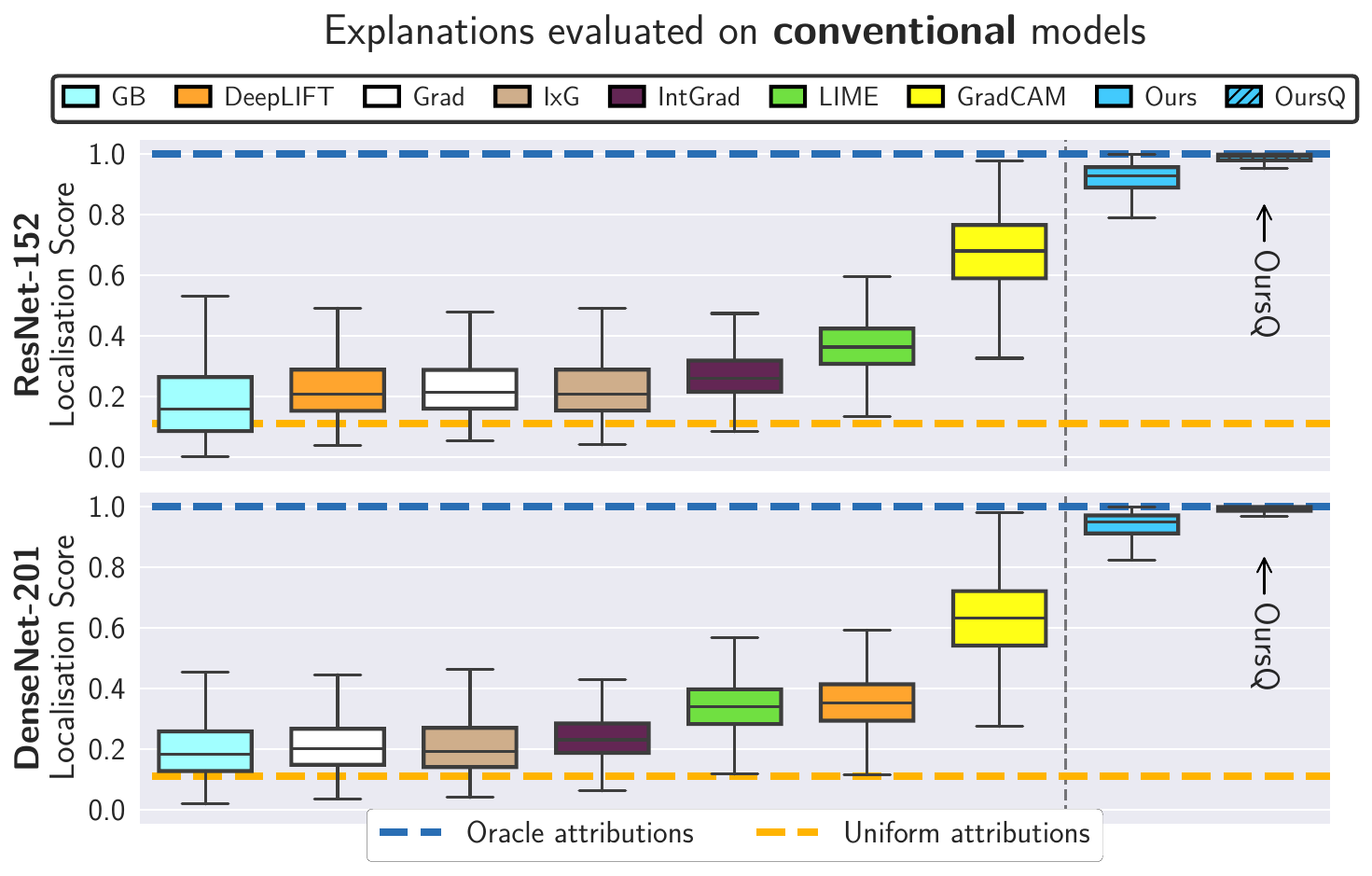}
    \end{subfigure}
    \caption{{Localisation results of the model-inherent contribution maps (`Ours', \cref{eq:contrib}) and various post-hoc explanation methods for a \bcos ResNet-152 and a \bcos DenseNet-201 (rows 1+2) and their conventional counterparts (rows 3+4); we further show the results of the top 0.025 percentile of pixels in the model-inherent explanations (`OursQ'), i.e.~roughly 11.3k pixels, 
    cf.~\cref{fig:localisation_results:quantiles}.
    For an easier cross-model comparison, we repeat the results of the \bcos models of rows 1+2 in rows 3+4 (Ours + OursQ). As can be seen, the model-inherent explanations significantly outperform post-hoc explanation methods when evaluating those methods on the \bcos models (rows 1+2) and on the conventional counterparts (rows 3+4). %
    }}
    \label{fig:localisation_results}
    \vspace{-1em}
\end{figure}

To provide more detail, in \cref{fig:localisation_results} we show the localisation results for two specific models (ResNet-152, DenseNet-201) for a wide range of post-hoc explanation methods: the `vanilla' gradient (Grad) \cite{baehrens2010explain}, Guided Backpropagation (GB) \cite{springenberg2014striving}, Input$\mytimes$Gradient (IxG) \cite{shrikumar2017deeplift}, Integrated Gradients (IntGrad) \cite{sundararajan2017axiomatic}, DeepLIFT \cite{shrikumar2017deeplift}, GradCAM \cite{selvaraju2017grad}, and LIME~\cite{ribeiro2016lime}. In particular, we plot the localisation scores for the \bcos versions of those models (rows 1+2) and the respective baselines (rows 3+4).

In comparison to the post-hoc explanation methods, we observe the model-inherent explanations (denoted as `Ours', \cref{eq:contrib}) to consistently and significantly outperform all other methods, both when comparing to post-hoc explanations for the \emph{same} model (rows 1+2), as well as when comparing explanations \emph{across} models (for ease of comparison, we repeat the \bcos results in rows 3+4). Note that while GradCAM also shows strong performance, it explains only a fraction of the entire model \cite{rao2022towards} and thus yields much coarser attribution maps, see also \cref{fig:quali_comp}. Interestingly, we also find that the other post-hoc explanations consistently perform better for \bcos models than for the baseline models (compare rows 1+3 and 2+4).

\begin{figure}
    \centering
    \hspace{-.65em}
    \begin{subfigure}[b]{\linewidth}
    \includegraphics[width=\textwidth]{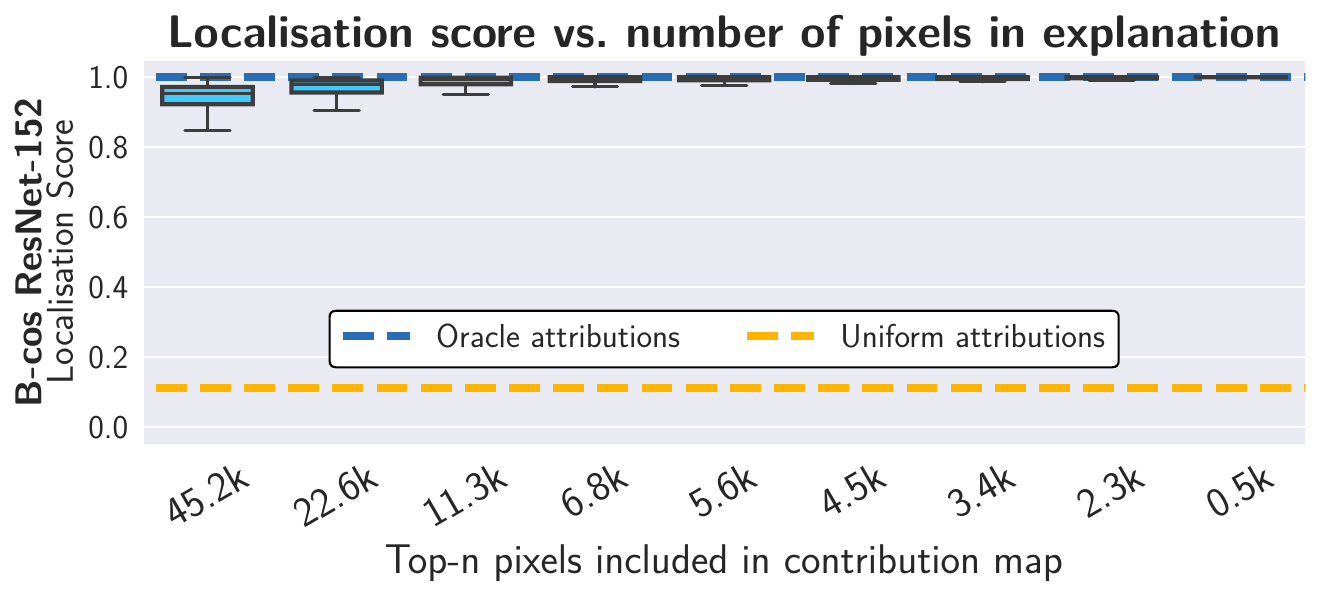}
    \end{subfigure}
    \caption{{\textbf{Top-n pixel localisation.} The localisation score significantly improves when only including the n most strongly contributing pixels of the model-inherent explanations in the localisation metric. While highlighting small contributions coming from images belonging to other classes (see \cref{fig:local_quali}) might be model-faithful, this is penalised in the metric. Nonetheless, we find that the correct region is consistently highlighted by the most strongly contributing pixels.
    }}
    \label{fig:localisation_results:quantiles}
\end{figure}

Finally, we additionally evaluate how well the most highly contributing pixels according to $\mat w(\vec x)$ fare in the localisation metric, which we denote as OursQuantile (OursQ) in \cref{fig:localisation_results}, see also \cref{fig:localisation_results:quantiles}. We find that when using only the top-n contributing pixels in the model-inherent explanations, the localisation score can be improved even further.%

\myparagraph{\bcos ViTs -- Accuracy.} In \cref{tbl:vit:90epochs}, we report the top-1 accuracies of \bcos ViT and ViT$_C$ models of various commonly used sizes---Tiny (Ti), Small (S), Base (B), and Large (L)---and the difference ($\Delta^3$) to the respective baseline models which we optimised under the same training protocol \cite{beyer2022better}.
\begin{table}[h!]
\centering
\addtolength{\tabcolsep}{-1pt}
\begin{tabular}{lcc|cccc}
 & \multicolumn{2}{c|}{\textbf{Accuracy}} & \multicolumn{4}{c}{\textbf{Localisation}} \\[.5em]
\textbf{\bcos Model}& \% & $\Delta^3$ & \%$\,$ & $\;\Delta_\text{Rollout}^\text{*}$ & $\Delta_\text{IntGrad}^\text{\bcos}$& $\Delta_\text{IntGrad}^\text{basel.}$ \\
\midrule
ViT-Ti &{\scriptsize  60.0} &{\scriptsize \color{red} -10.3} &{\scriptsize  69.8} &{\scriptsize \color{teal} +44.8} &{\scriptsize \color{teal} +12.2} &{\scriptsize \color{teal} +30.0} \\
ViT-S &{\scriptsize  69.2} &{\scriptsize \color{red} -5.2} &{\scriptsize  72.1} &{\scriptsize \color{teal} +47.1} &{\scriptsize \color{teal} +15.0} &{\scriptsize \color{teal} +30.9} \\
ViT-B &{\scriptsize  74.4} &{\scriptsize \color{red} -0.9} &{\scriptsize  74.6} &{\scriptsize \color{teal} +49.6} &{\scriptsize \color{teal} +19.4} &{\scriptsize \color{teal} +32.5} \\
ViT-L &{\scriptsize  75.1} &{\scriptsize \color{red} -0.7} &{\scriptsize  74.0} &{\scriptsize \color{teal} +49.0} &{\scriptsize \color{teal} +17.7} &{\scriptsize \color{teal} +32.2} \\
ViT$_{C}$-Ti &{\scriptsize  67.3} &{\scriptsize \color{red} -5.3} &{\scriptsize  86.8} &{\scriptsize \color{teal} +61.8} &{\scriptsize \color{teal} +30.3} &{\scriptsize \color{teal} +46.4} \\
ViT$_{C}$-S &{\scriptsize  74.5} &{\scriptsize \color{red} -1.2} &{\scriptsize  87.4} &{\scriptsize \color{teal} +62.4} &{\scriptsize \color{teal} +30.4} &{\scriptsize \color{teal} +45.9} \\
ViT$_{C}$-B &{\scriptsize  77.1} &{\scriptsize \color{teal} +0.3} &{\scriptsize  86.5} &{\scriptsize \color{teal} +61.5} &{\scriptsize \color{teal} +28.6} &{\scriptsize \color{teal} +44.5} \\
ViT$_{C}$-L &{\scriptsize  77.8} &{\scriptsize \color{red} -0.1} &{\scriptsize  87.3} &{\scriptsize \color{teal} +62.3} &{\scriptsize \color{teal} +29.9} &{\scriptsize \color{teal} +45.3} \\
\end{tabular}
\addtolength{\tabcolsep}{+1pt}
\caption{%
{
\textbf{Left:} Top-1 accuracies on ImageNet for the 90 epoch Simple-ViT training protocol \cite{beyer2022better}. $\Delta^3$: Accuracy difference with respect to training the baseline models according to the same protocol. While we observe a significant drop for smaller ViT and ViT$_C$ models, the difference to the baseline models vanishes for larger ViTs: e.g., the \bcos ViT$_C$-B and ViT$_C$-L models perform on par with the baselines. \textbf{Right:} Additionally, we show the localisation scores of the model-inherent explanations (\cref{eq:contrib}) as well as the differences to the scores obtained via Attention Rollout ($\Delta^*_\text{\scriptsize{Rollout}}$) and IntGrad evaluated on the \bcos models ($\Delta^\text{\scriptsize\bcos}_\text{\scriptsize IntG}$) and the baselines ($\Delta^\text{\scriptsize basel.}_\text{\scriptsize IntG}$). We observe significant localisation gains when using the \bcos explanations.}}
\label{tbl:vit:90epochs}
\end{table}

For both the original ViTs and the ones with a shallow convolutional stem (ViT$_C$), we observe significant performance drops for smaller model sizes (Ti+S). For larger models (B+L), however, the difference between the \bcos models and the baselines vanish, see, \eg, the ViT$_C$-B ($\Delta^3\myeq \myplus0.3$) and the ViT$_C$-L ($\Delta^3\myeq \myminus0.1$) models. Further, and in line with \cite{xiao2021early}, we find that a shallow convolutional stem leads to significant performance gains; as we discuss next, such a stem can also significantly improve model interpretability.

\myparagraph{\bcos ViTs -- Interpretability.} To assess the \bcos ViTs' interpretability, we compare the model-inherent explanations (\cref{eq:contrib}) to IntGrad evaluated on the \bcos ($\Delta^\text{\scriptsize \bcos}_\text{\scriptsize IntG}$) and the baseline models ($\Delta^\text{\scriptsize basel.}_\text{\scriptsize IntG}$); additionally, we report the difference to a commonly used attention-based explanation, namely Attention Rollout \cite{abnar2020quantifying} ($\Delta^\text{\scriptsize *}_\text{\scriptsize Rollout}$). Since Rollout is inherently a class-agnostic method\footnote{Attention Rollout computes the product of the average (across heads) attention matrices of all layers and is thus not class-specific.}, it on average yields the same localisation as uniformly distributed importance values for all models, \ie a mean localisation score of 25\%.

As can be seen in \cref{tbl:vit:90epochs}, the \bcos ViT and ViT$_C$ models show significant gains in localisation compared to the respective baseline models. Interestingly, adding a convolutional stem of just four convolutional layers yields notable improvements in localisation between the respective \bcos models, \ie between \bcos ViT and \bcos ViT$_C$ models.
This suggests that ViT$_C$ models perform better because the transformers are applied to more meaningful representations, which avoids mixing visually similar, yet semantically unrelated, patches in the first few global attention operations.

\begin{figure}
    \centering
    \begin{subfigure}[b]{\linewidth}\centering
    \includegraphics[width=\linewidth]{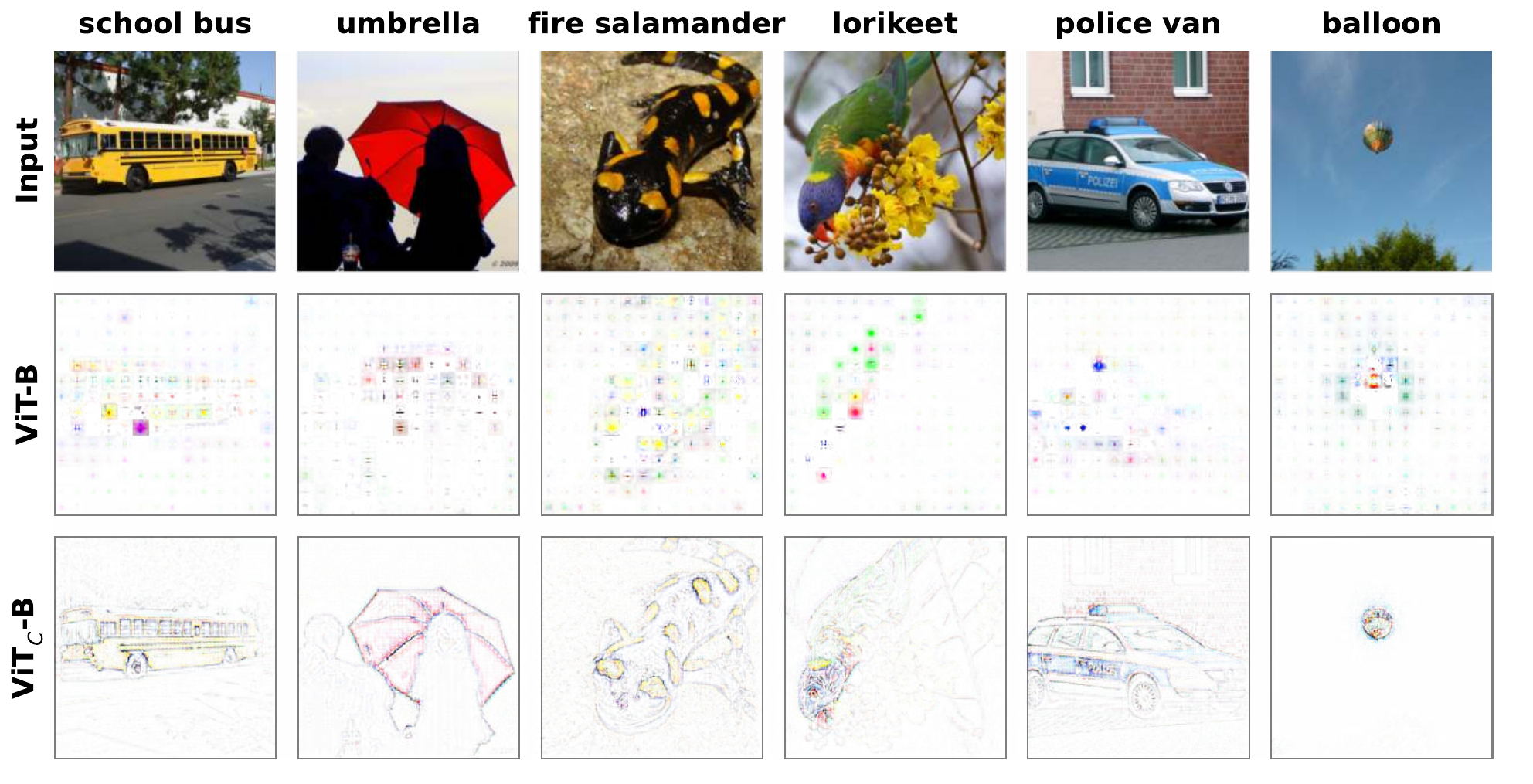}
    \end{subfigure}
    \begin{subfigure}[b]{\linewidth}\centering
    \includegraphics[width=\linewidth]{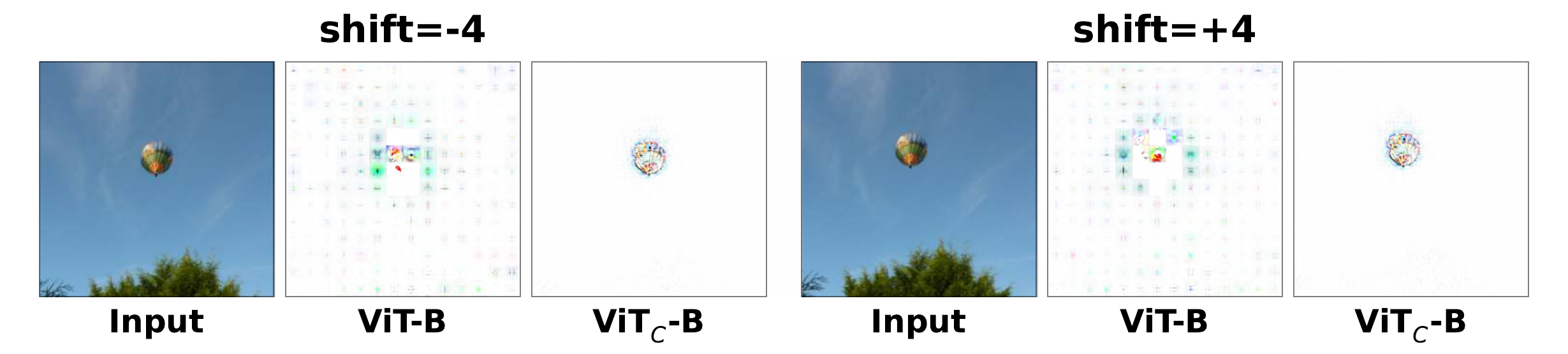}
    \end{subfigure}
    \caption{{\textbf{Top:} Comparison between explanations of \bcos ViT-B models without (\textbf{row 2}) and with (\textbf{row 3}) convolutional stem, \textit{i}.\textit{e}. between ViT and ViT$_C$ models. By adding just 4 convolutional layers for embedding the image patches, the explanations become much more coherent and less sparse. \textbf{Bottom:} For a single image, we qualitative visualise how the explanations of the different ViT models perform under diagonal shifts $s\myin{-4, +4}$ of the image on the top right. While both models generally show consistent explanations, the inductive bias due to the convolutional stem leads to more stable behaviour and explanations of the ViT$_C$ models. %
    }}
    \label{fig:quali:vits}
\end{figure}

\begin{figure}[t!]
    \centering
    \begin{subfigure}[b]{\linewidth}
    \includegraphics[width=\linewidth, trim=.25em 1em .5em 0, clip]{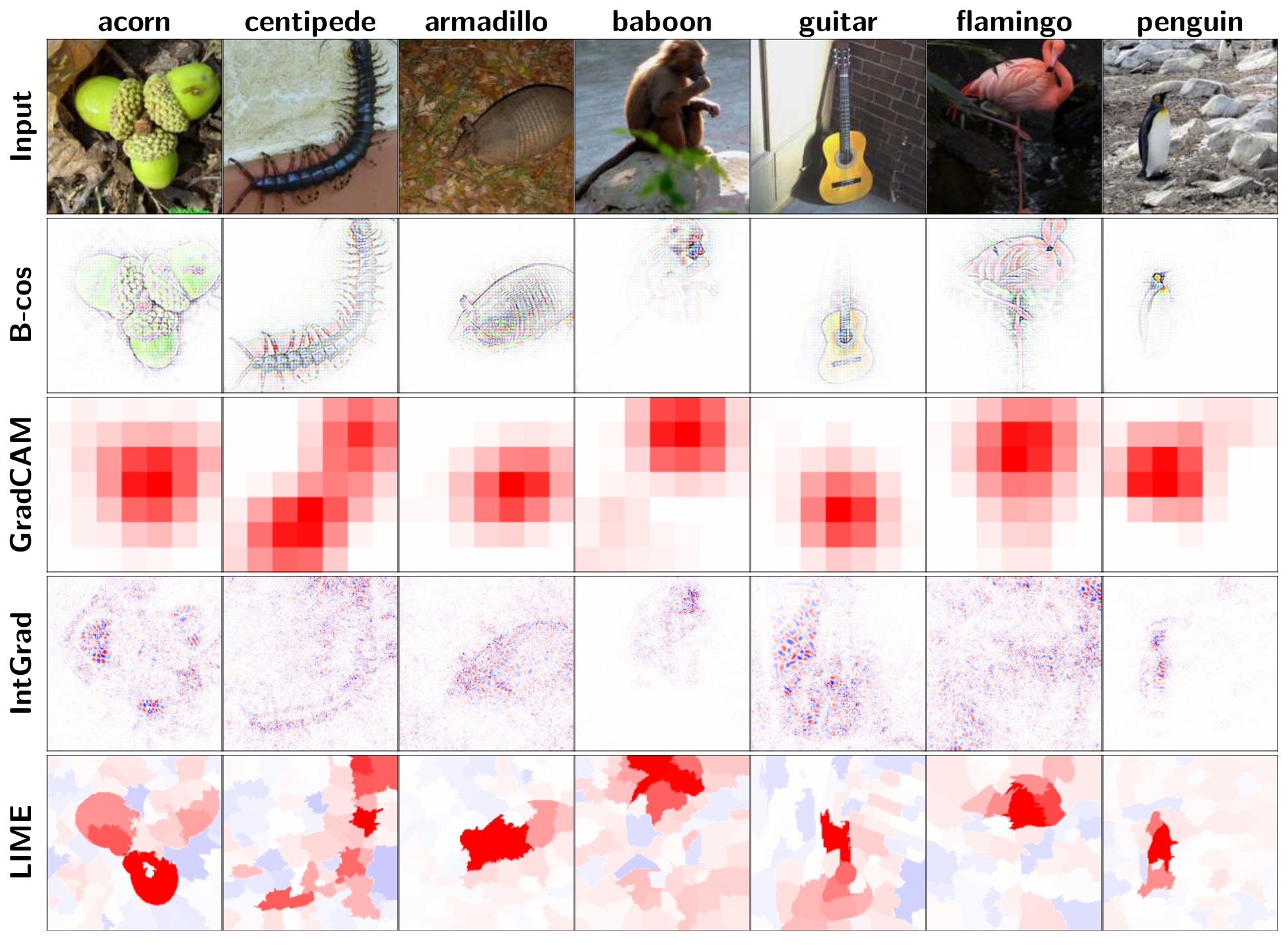}
    \end{subfigure}
    \caption{{Comparison between \bcos explanations and various commonly used post-hoc methods on a \bcos DenseNet-121. For all post-hoc explanations, positive, negative, and zero attributions are shown in red, blue, and white respectively.}}
    \label{fig:quali_comp}
\end{figure}

These \emph{quantitative} improvements in interpretability between ViT and ViT$_C$ models can also be observed qualitatively, see \cref{fig:quali:vits}. Specifically, in contrast to the explanations of the ViTs without convolutional stem (row 2), the \bcos ViT$_C$ explanations are visually more coherent and less sparse, making them more easily interpretable for humans.

\subsection{Qualitative evaluation of explanations}
\label{subsec:quali_results}
{The following results are based on the DenseNet-121, cf.~\cref{tbl:cnn:90epochs}, we found similar results for other \bcos models.} 

Every activation in a \bcos model is the result of a sequence of \bcos transforms. Hence, the intermediate activations in any layer $l$ can also be explained via the corresponding linear transform $\mat W_{1\rightarrow l}(\vec x)$, see \cref{eq:collapse}. 

For example, in \cref{fig:global_protos}, we visualise the linear transforms of the respective \emph{class logits} for various input images.
Given the alignment pressure during optimisation, these linear transforms align with class-discriminant patterns in the input and thus actually resemble the class objects. 

\begin{figure}[t!]
    \centering
    \vspace{-.2em}
    \hspace{-.6em}
    \begin{subfigure}[b]{.975\linewidth}
    \begin{subfigure}[b]{\linewidth}
    \includegraphics[width=\textwidth]{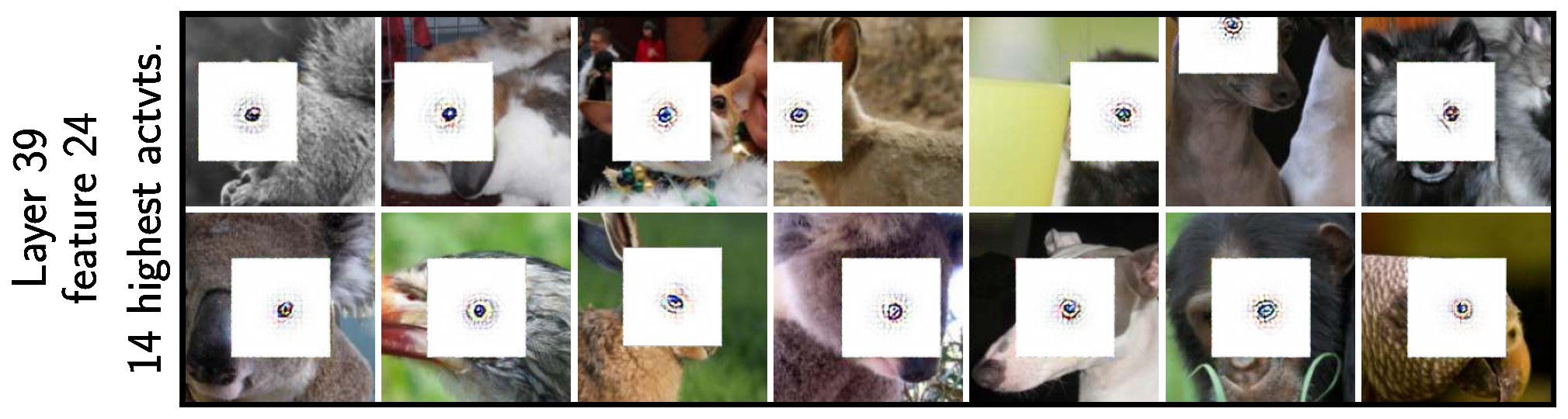}
    \end{subfigure}
    \begin{subfigure}[b]{\linewidth}
    \includegraphics[width=\textwidth]{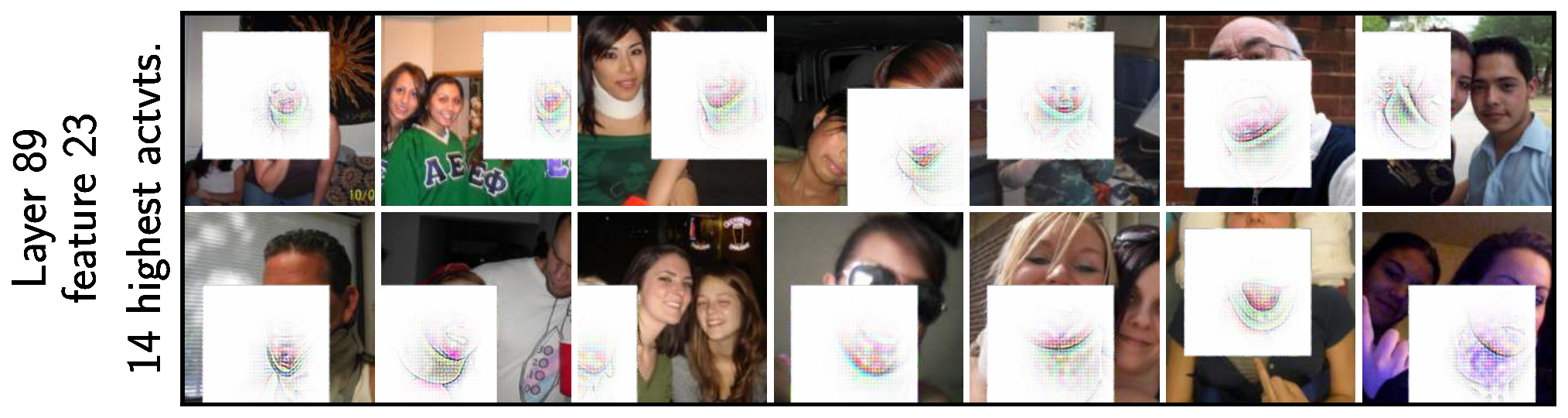}
    \end{subfigure}
    \begin{subfigure}[b]{\linewidth}
    \includegraphics[width=\textwidth]{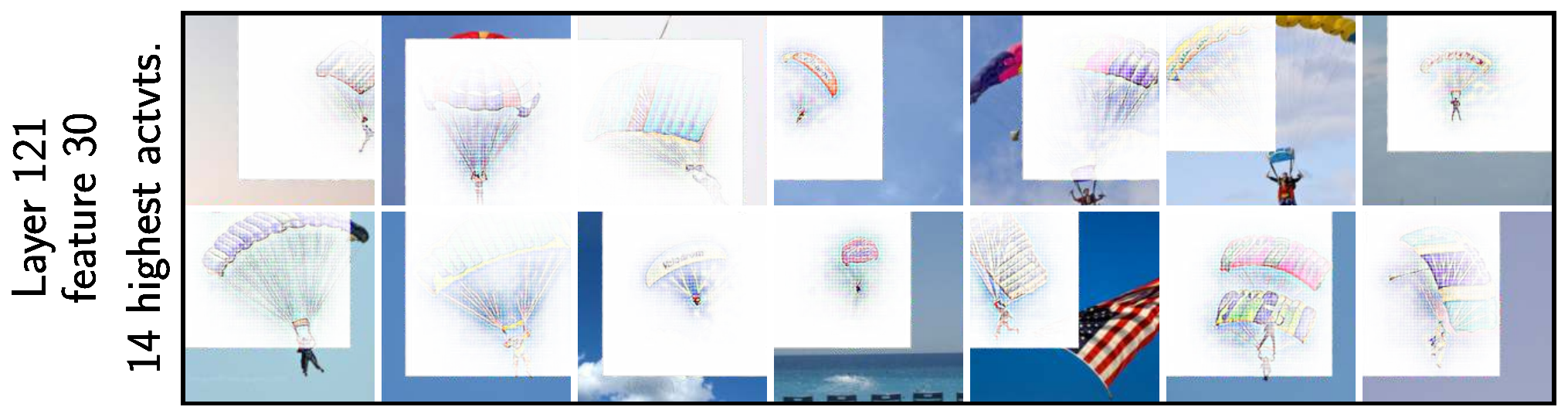}
    \end{subfigure}
    \end{subfigure}
    \caption{Explanations for highly contributing feature directions in various layers. In early layers, the features seem to encode low-level concepts (e.g., eyes in layer 39) and represent more high-level concepts in later layers (e.g., layers 89 and 121), such as neck braces or parachutes, see also \cref{fig:intermediate_protos}.}
    \label{fig:other_layers}
    \vspace{-.2em}
\end{figure}

Similarly, in \cref{fig:intermediate_protos,,fig:other_layers}, we visualise explanations for \emph{intermediate features}. These features are selected from the most highly contributing feature directions $\hat{\vec a}^l_{(x, y)}\myin \mathbb R^{d_l}$ according to the linear transformation $\mat W_{l\rightarrow L}(\vec x)$, cf.~\cref{eq:collapse}; here, $\hat{\vec a}$ denotes that $\hat{\vec a}$ is of unit norm, $d_l$ is the number of features in layer $l$, and $(x, y)$ denotes the position at which this feature is extracted from the activation map.

Specifically, in \cref{fig:other_layers}, we show explanations for some images that most strongly activate those feature directions across the validation set. We find that features in early layers seem to represent low-level concepts, and become more complex in later layers.
\begin{figure}[b!]
    \centering
    \begin{subfigure}[b]{.9\linewidth}
    \begin{subfigure}[b]{\linewidth}
    \includegraphics[width=\linewidth, ]{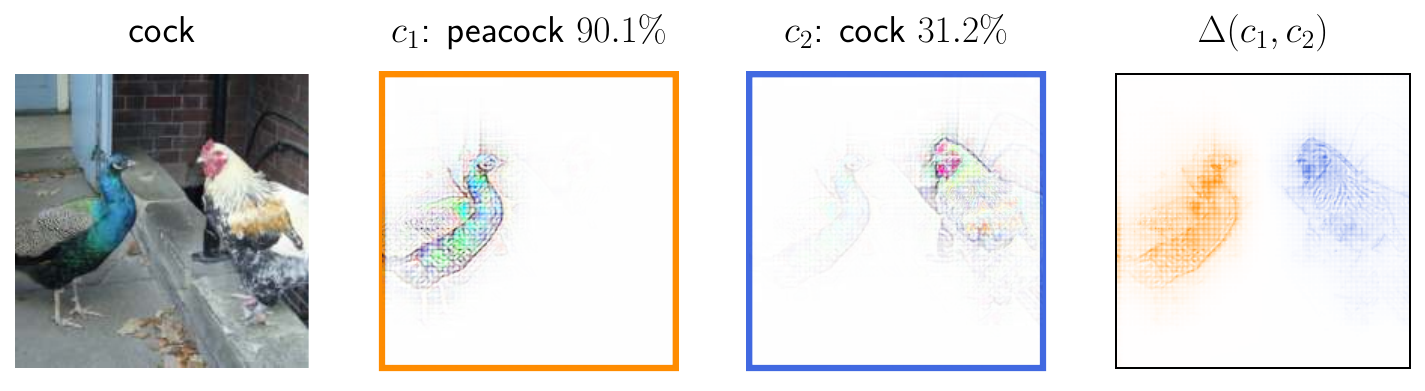}
    \end{subfigure}
    \begin{subfigure}[b]{\linewidth}
    \includegraphics[width=\linewidth, ]{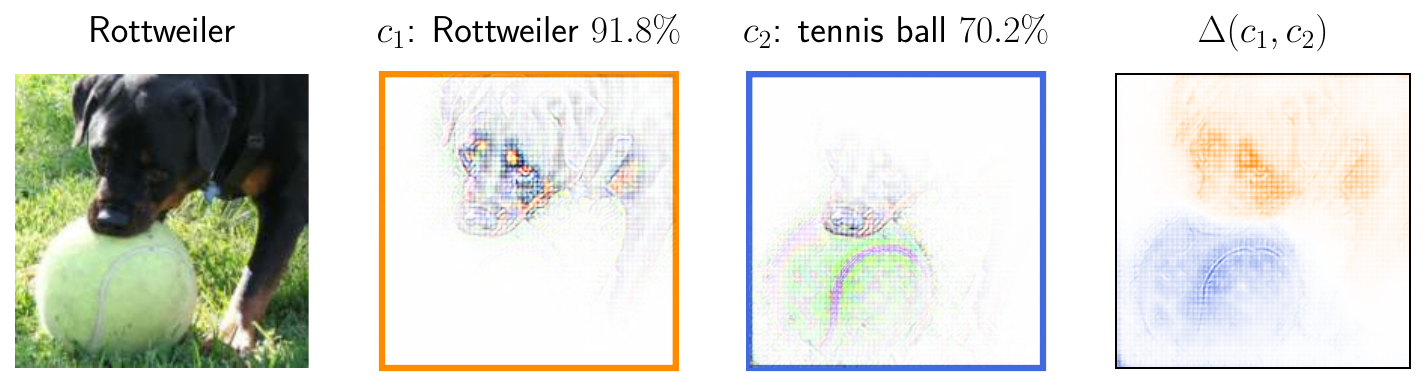}
    \end{subfigure}
    \begin{subfigure}[b]{\linewidth}
    \includegraphics[width=\linewidth, ]{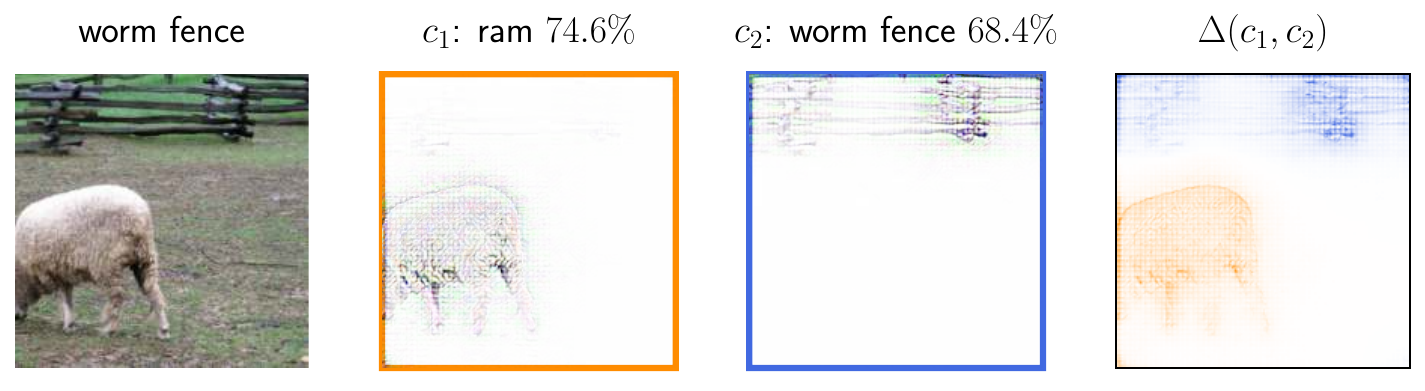}
    \end{subfigure}
    \end{subfigure}
    \caption{\hspace{-.15em}\textbf{Col.~1:} Ambiguous input image. \textbf{Cols.~2+3:} Explanations for the two most likely classes under a \bcos DenseNet-121. \textbf{Col.~4:} Difference of contribution maps for the two class logits, i.e., $\Delta(c_1, c_2)\myeq\vec s_{c_1}^L(\vec x)\myminus\vec s_{c_2}^L(\vec x)$, see \cref{eq:contrib}; positive values shown in orange ($c_1$), negative values in blue ($c_2$).}
    \label{fig:ambiguous_classifications}
    \vspace{-.75em}
\end{figure}
\begin{figure*}[t!]
    \vspace{-.75em}
    \centering
    \begin{subfigure}[b]{.975\textwidth}
    \centering
    \begin{subfigure}[b]{\linewidth}
    \includegraphics[width=\textwidth]{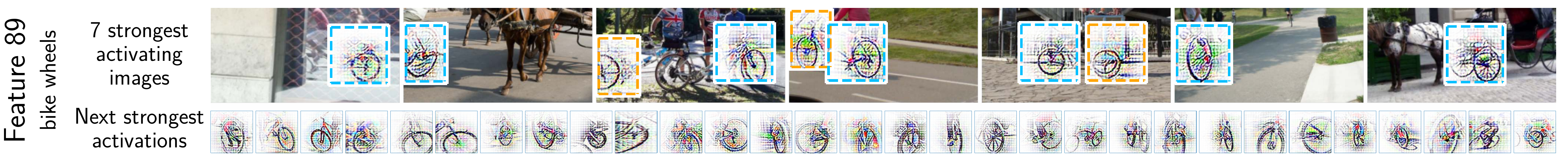}
    \end{subfigure}
    \begin{subfigure}[b]{\linewidth}
    \includegraphics[width=\textwidth]{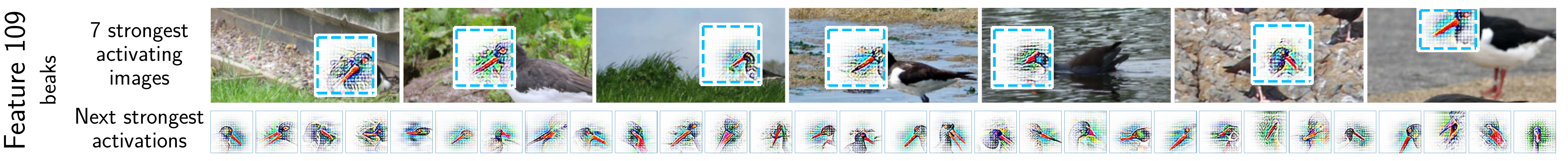}
    \end{subfigure}
    \begin{subfigure}[b]{\linewidth}
    \includegraphics[width=\textwidth]{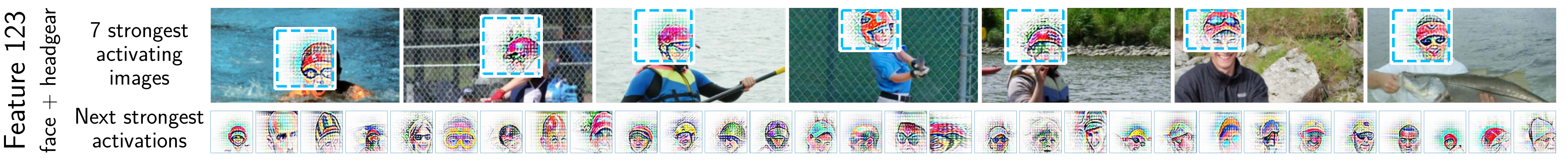}
    \end{subfigure}
    \begin{subfigure}[b]{\linewidth}
    \includegraphics[width=\textwidth]{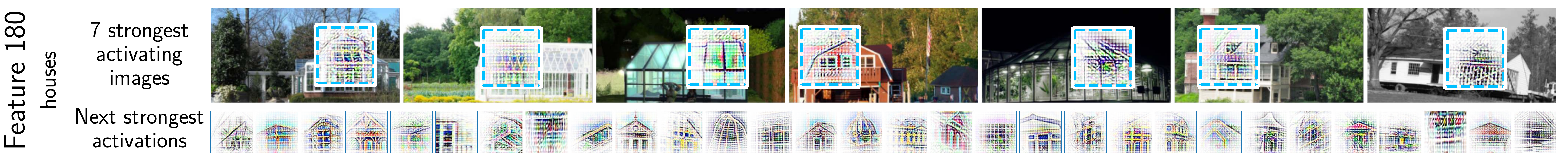}
    \end{subfigure}
    \begin{subfigure}[b]{\linewidth}
    \includegraphics[width=\textwidth]{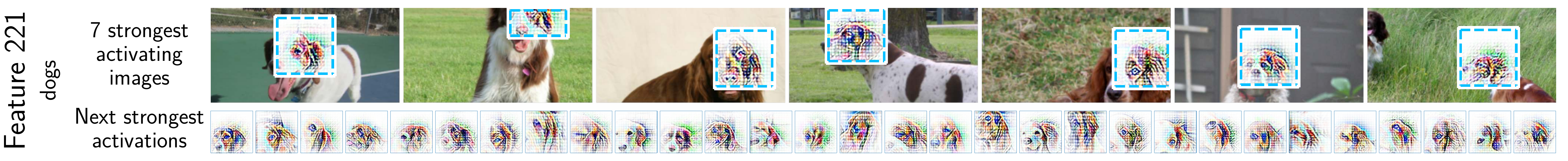}
    \end{subfigure}
    \end{subfigure}
    \caption[Caption for LOF]{
    Explanations of 5 highly contributing feature directions in layer 87 of a DenseNet-121.
    For each direction, we provide its index number $n$ and a concept description.
    Further, we show the 7 most activating images for each direction (top row per feature), in which we visualise the explanation for the highest (\textbf{blue squares}) activation; i.e., visualise the 72$\times$72 center patch of the weighting $[\mat w_{1\rightarrow l}(\vec x)]_n$ for feature $n$. For some images in the first row, we additionally show the explanation for the 2nd highest activation (\textbf{orange squares}). Lastly, we show the explanations of the highest activations (corresponding to the blue squares) for the next 30 images to highlight the features' specificity.
    }
    \label{fig:intermediate_protos}
    \vspace{-1.25em}
\end{figure*}

\cref{fig:intermediate_protos} shows additional results for features in layer 87. We observed that some features are highly specific to certain concepts, such as bike wheels (feature 89), red bird beaks (feature 109), or faces with headgear (feature 123). Importantly, these features do not just learn to align with simple, fixed patterns---instead, they represent semantic concepts and are robust to changes in colour, size, and pose.

Lastly, 
in \cref{fig:ambiguous_classifications}, we show explanations of the two most likely classes for images for which the model produces predictions with high uncertainty; additionally, we show the $\Delta$-Explanation, i.e., the difference in contribution maps for the two classes, see \cref{eq:contrib}. By means of the model-inherent linear mappings $\mat w_{1\rightarrow L}$, the model can provide a human-interpretable explanation for its uncertainty: there are indeed features in each of those images that provide evidence for both of the predicted classes.

\subsection{Explicit and implicit model biases}
\label{subsec:biases}
As discussed in \cref{subsubsec:normed_bcos}, we design the \bcos CNNs to be bias-free by fixing the \emph{explicit} bias parameters in all layers to be zero. In this section, we aim to motivate this choice further. Additionally, we show that even when doing so, the models can learn to counteract this and add \emph{implicit} biases by exploiting reliable input features such as image edges.

\begin{figure}
    \centering
    \begin{subfigure}[b]{\linewidth}
    \includegraphics[width=\linewidth]{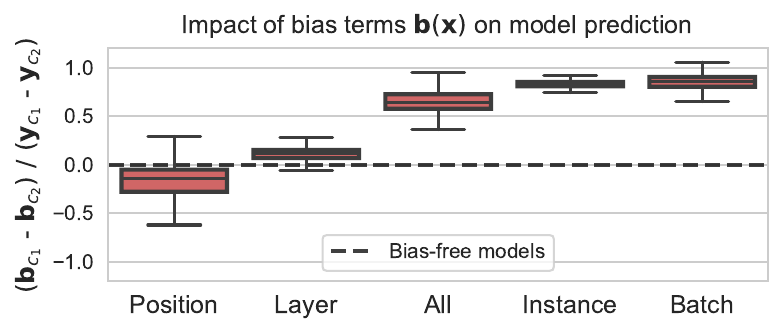}
    \end{subfigure}
    \caption{{\textbf{Ratio of bias differences to logit differences} for the top-2 predictions ($c_1$, $c_2$) of ResNet-20 models on the CIFAR10 test set, trained with different normalisation layers. For some models, the bias term explains most of the difference and thus plays a critical role in the model predictions (e.g., for BatchNorm and InstanceNorm). On the other hand, for the model trained with LayerNorm, the bias term does not seem to be decisive, whereas for PositionNorm it even seems to favour the second highest prediction. To avoid not adequately representing the potentially critical bias $\vec b(\vec x)$ in the explanations, we ensure $\vec b(\vec x)\myeq0$ for \bcos CNNs.} }
    \label{fig:norm-ablation}
\end{figure}

\myparagraph{Explicit biases.} As discussed by \cite{srinivas2019full} for piece-wise linear networks, the biases of DNNs are often not accounted for in input-level attributions, despite the fact that they can play a critical role in the model prediction. Similarly, the contributions $\vec s_c$ (\cref{eq:contrib}) from the input to the $c$-th class logit are not complete explanations if biases are used within \bcos models {(cf.~{\cref{subsubsec:normed_bcos}})}. Instead, the output $\vec y(\vec x)$ is
\begin{align}
    \vec y (\vec x) = \mat W(\vec x) \vec x + \vec b(\vec x)\quad,
\end{align}
with $\vec b(\vec x)\myin \mathbb R^{c}$ subsuming all the contributions to the class logits that are not accounted for by the dynamic linear mapping $\mat w(\vec x)$. As such, \bcos DNNs face the same problem as described by \cite{srinivas2019full}: while explanations based on $\mat W(\vec x)$ neglect the impact of the contributions from bias terms $\vec b(\vec x)$, in 
\cref{fig:norm-ablation} we show that the bias term can play a critical role in the model decision. In particular, we find that for some models (\eg see the models trained with InstanceNorm or BatchNorm), the bias terms make up most of the difference between the top-2 predictions of the model. To ensure that the explanations of \bcos CNNs are complete, we set all bias parameters of the model to zero, see \cref{subsubsec:normed_bcos}. 

\begin{figure}
    \centering
    \begin{subfigure}[b]{\linewidth}
    \includegraphics[width=\linewidth]{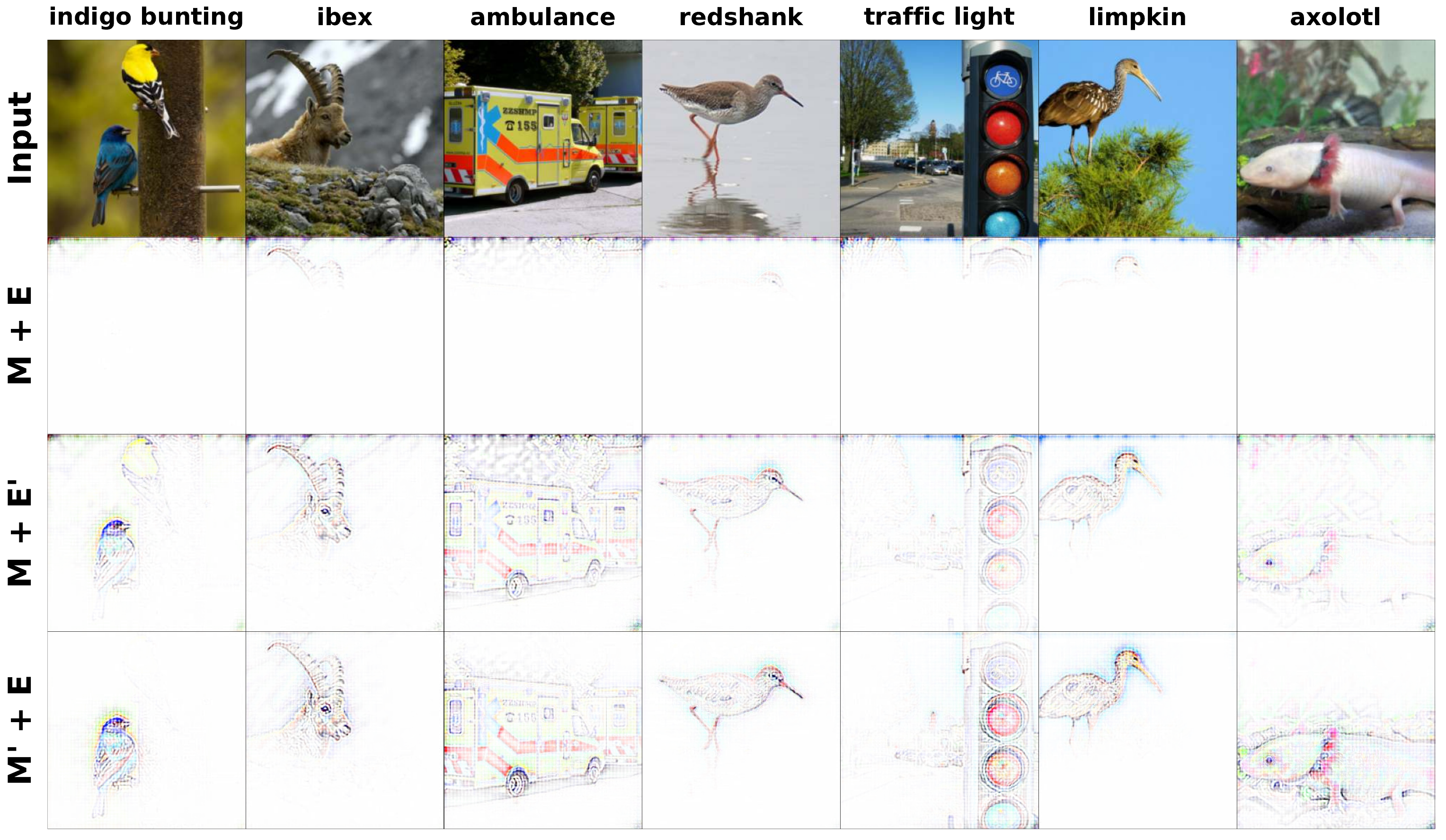}
    \end{subfigure}
    \caption{{\textbf{`You get what you optimise for.'}
    We found that for some models M (here, ResNet-18), the explanations E seem to highlight image edges for all samples, implying that most \emph{positive} contributions for all classes come from those edges,  see \textbf{row 2}. 
    However, the explanations for the difference from the mean logit (i.e., $\mathbf{y}_c \myminus \langle\mathbf y_j\rangle_j$)---denoted by E'---are still qualitatively convincing, see \textbf{row 3}, suggesting that the models learn to add a positive bias to all classes, which vanishes in the mean-corrected explanations. To alleviate this, we propose to optimise the models differently (yielding M', see \textbf{row 4}) instead of changing the explanations: specifically, by changing the target encoding, as described in \cref{subsubsec:optim}, we encourage the models to focus on positive evidence.
    }}
    \label{fig:ablation:top-bias}
\end{figure}

\myparagraph{Implicit biases.}
Even when no \emph{explicit} bias parameters are used, we found that some models learnt to use image edges for computing biases to add to the class logits. While these biases are correctly reflected in the explanations (see \cref{fig:ablation:top-bias}, row 2), this behaviour makes the model explanations less human-interpretable. Interestingly, these models seem to have learnt to solve the optimisation task in \cref{eq:loss} in an unintended manner. Specifically, all classes receive highly positive contributions from image corners or edges (the top edge in \cref{fig:ablation:top-bias}), thus yielding high class logits for all classes. To still obtain low scores for the non-target classes, the models then seem to use features of the recognised object in the image to add negative contributions to those classes. 

To corroborate this, we additionally visualise the explanations for the mean-corrected logits $\vec y \myminus \langle \vec y_c\rangle_c$ in row 3 of \cref{fig:ablation:top-bias}; note that given the dynamic linearity of the \bcos models, this is equivalent to computing mean-corrected explanations $\mat W(\vec x) \myminus \langle[\mat W(\vec x)]_c\rangle_c$. As expected, we find that those modified explanations E' exhibit a high degree of specificity and are thus more easily human-interpretable.

We hypothesise that this behaviour is rooted in the formulation of the optimisation task itself: in particular, the BCE loss 'asks' the model to produce highly negative logits for all classes but one, which might bias the model towards using object features to compute negative contributions.

To overcome this, and to incentivise the models to use object features primarily to compute \emph{positive contributions} for the classes this feature belongs to, we propose to change the optimisation task itself, as described in \cref{subsubsec:optim}. As we show in the last row of \cref{fig:ablation:top-bias}, this indeed has the intended effect and results in explanations that are highly focused on the class objects in the image. 

We believe this to highlight the importance of a holistic approach towards interpretable deep neural networks. 
In particular, these results show that both the model design as well as the optimisation procedure can significantly impact the model behaviour as well as the resulting explanations, which complicates the development of post-hoc explanations methods that do not take those aspects into account.

\color[RGB]{0, 0, 0}
\section{Conclusion}
\label{sec:discussion}
We presented a novel approach for endowing deep neural networks with a high degree of \emph{inherent interpretability}. 
In particular, we developed the \bcos transformation as a modification of the linear transformation to increase weight-input alignment during optimisation and showed that this can significantly increase interpretability. 
Importantly, the \bcos transformations can be used as a drop-in replacement for the ubiquitously used linear transformations in conventional DNNs whilst only incurring minor drops in classification accuracy. 
As such, our approach can increase the interpretability of a wide range of DNNs at a low cost and thus holds great potential to have a significant impact on the deep learning community. In particular, it shows that strong performance and interpretability need not be at odds. Moreover, we demonstrate that by structurally constraining \emph{how} the neural networks are to solve an optimisation task---in the case of \bcos networks via \emph{alignment}---allows for extracting explanations that faithfully reflect the underlying model.
We believe this to be  an important step on the road towards interpretable deep learning, which is an essential ingredient for building trust in DNN-based decisions, especially in safety-critical situations.

\appendices

\bibliographystyle{IEEEtran}
\bibliography{bibliography}

\begin{thebibliography}{10}
\providecommand{\url}[1]{#1}
\csname url@samestyle\endcsname
\providecommand{\newblock}{\relax}
\providecommand{\bibinfo}[2]{#2}
\providecommand{\BIBentrySTDinterwordspacing}{\spaceskip=0pt\relax}
\providecommand{\BIBentryALTinterwordstretchfactor}{4}
\providecommand{\BIBentryALTinterwordspacing}{\spaceskip=\fontdimen2\font plus
\BIBentryALTinterwordstretchfactor\fontdimen3\font minus
  \fontdimen4\font\relax}
\providecommand{\BIBforeignlanguage}[2]{{%
\expandafter\ifx\csname l@#1\endcsname\relax
\typeout{** WARNING: IEEEtran.bst: No hyphenation pattern has been}%
\typeout{** loaded for the language `#1'. Using the pattern for}%
\typeout{** the default language instead.}%
\else
\language=\csname l@#1\endcsname
\fi
#2}}
\providecommand{\BIBdecl}{\relax}
\BIBdecl

\bibitem{xai_overview}
W.~Samek, G.~Montavon, S.~Lapuschkin, C.~J. Anders, and K.-R. Müller,
  ``{Explaining Deep Neural Networks and Beyond: A Review of Methods and
  Applications},'' \emph{Proceedings of the IEEE}, vol. 109, no.~3, 2021.

\bibitem{nair2010relu}
V.~Nair and G.~E. Hinton, ``{Rectified Linear Units Improve Restricted
  Boltzmann Machines},'' in \emph{{International Conference on Machine Learning
  (ICML)}}, 2010.

\bibitem{montufar2014number}
G.~F. Mont{\'{u}}far, R.~Pascanu, K.~Cho, and Y.~Bengio, ``{On the Number of
  Linear Regions of Deep Neural Networks},'' in \emph{{Advances in Neural
  Information Processing Systems (NeurIPS)}}, 2014.

\bibitem{BAM2019}
M.~Yang and B.~Kim, ``{Benchmarking Attribution Methods with Relative Feature
  Importance},'' \emph{CoRR}, vol. abs/1907.09701, 2019.

\bibitem{shrikumar2017deeplift}
A.~Shrikumar, P.~Greenside, and A.~Kundaje, ``{Learning Important Features
  Through Propagating Activation Differences},'' in \emph{{International
  Conference on Machine Learning (ICML)}}, 2017.

\bibitem{adebayo2018sanity}
J.~Adebayo, J.~Gilmer, M.~Muelly, I.~J. Goodfellow, M.~Hardt, and B.~Kim,
  ``{Sanity Checks for Saliency Maps},'' in \emph{{Advances in Neural
  Information Processing Systems (NeurIPS)}}, 2018.

\bibitem{krizhevsky2009cifar10}
A.~Krizhevsky, ``{Learning multiple layers of features from tiny images},''
  University of Toronto, Tech. Rep., 2009.

\bibitem{dosovitskiy2021an}
A.~Dosovitskiy, L.~Beyer, A.~Kolesnikov, D.~Weissenborn, X.~Zhai,
  T.~Unterthiner, M.~Dehghani, M.~Minderer, G.~Heigold, S.~Gelly, J.~Uszkoreit,
  and N.~Houlsby, ``{An Image is Worth 16x16 Words: Transformers for Image
  Recognition at Scale},'' in \emph{{International Conference on Learning
  Representations (ICLR)}}, 2021.

\bibitem{simonyan2015vgg}
K.~Simonyan and A.~Zisserman, ``{Very Deep Convolutional Networks for
  Large-Scale Image Recognition},'' in \emph{{International Conference on
  Learning Representations (ICLR)}}, 2015.

\bibitem{he2016deep}
K.~He, X.~Zhang, S.~Ren, and J.~Sun, ``{Deep Residual Learning for Image
  Recognition},'' in \emph{{Proceedings of the IEEE Conference on Computer
  Vision and Pattern Recognition (CVPR)}}, 2016.

\bibitem{huang2017densely}
G.~Huang, Z.~Liu, L.~Van Der~Maaten, and K.~Q. Weinberger, ``{Densely Connected
  Convolutional Networks},'' in \emph{{Proceedings of the IEEE Conference on
  Computer Vision and Pattern Recognition (CVPR)}}, 2017.

\bibitem{xie2017aggregated}
S.~Xie, R.~Girshick, P.~Doll{\'a}r, Z.~Tu, and K.~He, ``{Aggregated residual
  transformations for deep neural networks},'' in \emph{{Proceedings of the
  IEEE conference on computer vision and pattern recognition}}, 2017.

\bibitem{liu2022convnet}
Z.~Liu, H.~Mao, C.-Y. Wu, C.~Feichtenhofer, T.~Darrell, and S.~Xie, ``{A
  convnet for the 2020s},'' in \emph{{Proceedings of the IEEE/CVF Conference on
  Computer Vision and Pattern Recognition}}, 2022.

\bibitem{Boehle2022CVPR}
M.~Böhle, M.~Fritz, and B.~Schiele, ``{B-cos Networks: Alignment is All we
  Need for Interpretability},'' in \emph{IEEE/CVF Conference on Computer Vision
  and Pattern Recognition ({CVPR})}, 2022.

\bibitem{lundberg2017unified}
S.~M. Lundberg and S.~Lee, ``{A Unified Approach to Interpreting Model
  Predictions},'' in \emph{{Advances in Neural Information Processing Systems
  (NeurIPS)}}, 2017.

\bibitem{petsiuk2018rise}
V.~Petsiuk, A.~Das, and K.~Saenko, ``{{RISE:} Randomized Input Sampling for
  Explanation of Black-box Models},'' in \emph{{British Machine Vision
  Conference (BMVC)}}, 2018.

\bibitem{ribeiro2016lime}
M.~T. Ribeiro, S.~Singh, and C.~Guestrin, ``{"{W}hy {S}hould {I} {T}rust
  {Y}ou?": Explaining the Predictions of Any Classifier},'' in \emph{{
  International Conference on Knowledge Discovery and Data Mining (SIGKDD)}},
  2016.

\bibitem{kim2018tcav}
B.~Kim, M.~Wattenberg, J.~Gilmer, C.~J. Cai, J.~Wexler, F.~B. Vi{\'{e}}gas, and
  R.~Sayres, ``{Interpretability Beyond Feature Attribution: Quantitative
  Testing with Concept Activation Vectors {(TCAV)}},'' in \emph{{International
  Conference on Machine Learning (ICML)}}, 2018.

\bibitem{das2020prototypes}
S.~Das, P.~Xu, Z.~Dai, A.~Endert, and L.~Ren, ``{Interpreting Deep Neural
  Networks through Prototype Factorization},'' in \emph{{International
  Conference on Data Mining Workshops (ICDMW)}}, 2020.

\bibitem{simonyan2013deep}
K.~Simonyan, A.~Vedaldi, and A.~Zisserman, ``{Deep Inside Convolutional
  Networks: Visualising Image Classification Models and Saliency Maps},'' in
  \emph{{International Conference on Learning Representations (ICLR),
  Workshop}}, 2014.

\bibitem{springenberg2014striving}
J.~T. Springenberg, A.~Dosovitskiy, T.~Brox, and M.~A. Riedmiller, ``{Striving
  for Simplicity: The All Convolutional Net},'' in \emph{{International
  Conference on Learning Representations (ICLR), Workshop}}, 2015.

\bibitem{zhou2016CAM}
B.~Zhou, A.~Khosla, {\`{A}}.~Lapedriza, A.~Oliva, and A.~Torralba, ``{Learning
  Deep Features for Discriminative Localization},'' in \emph{{Proceedings of
  the IEEE Conference on Computer Vision and Pattern Recognition (CVPR)}},
  2016.

\bibitem{bach2015pixel}
S.~Bach, A.~Binder, G.~Montavon, F.~Klauschen, K.-R. M{\"u}ller, and W.~Samek,
  ``{On Pixel-Wise Explanations for Non-Linear Classifier Decisions by
  Layer-Wise Relevance Propagation},'' \emph{{PLoS ONE}}, 2015.

\bibitem{selvaraju2017grad}
R.~R. Selvaraju, M.~Cogswell, A.~Das, R.~Vedantam, D.~Parikh, and D.~Batra,
  ``{Grad-CAM: Visual Explanations from Deep Networks via Gradient-Based
  Localization},'' in \emph{{International Conference on Computer Vision
  (ICCV)}}, 2017.

\bibitem{srinivas2019full}
S.~Srinivas and F.~Fleuret, ``{Full-Gradient Representation for Neural Network
  Visualization},'' in \emph{{Advances in Neural Information Processing Systems
  (NeurIPS)}}, 2019.

\bibitem{sundararajan2017axiomatic}
M.~Sundararajan, A.~Taly, and Q.~Yan, ``{Axiomatic Attribution for Deep
  Networks},'' in \emph{{International Conference on Machine Learning (ICML)}},
  2017.

\bibitem{chen2019looks}
C.~Chen, O.~Li, D.~Tao, A.~Barnett, C.~Rudin, and J.~Su, ``{This Looks Like
  That: Deep Learning for Interpretable Image Recognition},'' in
  \emph{{Advances in Neural Information Processing Systems (NeurIPS)}}, 2019.

\bibitem{brendel2018approximating}
W.~Brendel and M.~Bethge, ``{Approximating CNNs with Bag-of-local-Features
  models works surprisingly well on ImageNet},'' in \emph{{International
  Conference on Learning Representations (ICLR)}}, 2019.

\bibitem{Boehle2021CVPR}
M.~Böhle, M.~Fritz, and B.~Schiele, ``{Convolutional Dynamic Alignment
  Networks for Interpretable Classifications},'' in \emph{{Proceedings of the
  IEEE Conference on Computer Vision and Pattern Recognition (CVPR)}}, 2021.

\bibitem{tsipras2019atodds}
D.~Tsipras, S.~Santurkar, L.~Engstrom, A.~Turner, and A.~Madry, ``{Robustness
  May Be at Odds with Accuracy},'' in \emph{{International Conference on
  Learning Representations (ICLR)}}, 2019.

\bibitem{Shah2021grads}
H.~Shah, P.~Jain, and P.~Netrapalli, ``{Do Input Gradients Highlight
  Discriminative Features?}'' \emph{CoRR}, vol. abs/2102.12781, 2021.

\bibitem{kim2019safeml}
B.~Kim, J.~Seo, and T.~Jeon, ``{Bridging Adversarial Robustness and Gradient
  Interpretability},'' \emph{{Safe Machine Learning workshop at International
  Conference on Learning Representations (ICLR)}}, 2019.

\bibitem{srinivas2021rethink}
S.~Srinivas and F.~Fleuret, ``{Rethinking the Role of Gradient-based
  Attribution Methods for Model Interpretability},'' in \emph{{International
  Conference on Learning Representations (ICLR)}}, 2021.

\bibitem{kindermans2018learning}
P.-J. Kindermans, K.~T. Schütt, M.~Alber, K.-R. Müller, D.~Erhan, B.~Kim, and
  S.~Dähne, ``{Learning how to explain neural networks: PatternNet and
  PatternAttribution},'' in \emph{{International Conference on Learning
  Representations (ICLR)}}, 2018.

\bibitem{zoumpourlis2017non}
G.~Zoumpourlis, A.~Doumanoglou, N.~Vretos, and P.~Daras, ``{Non-linear
  convolution filters for CNN-based learning},'' in \emph{{Proceedings of the
  IEEE Conference on Computer Vision and Pattern Recognition (CVPR)}}, 2017.

\bibitem{liu2017hyperspherical}
W.~Liu, Y.~Zhang, X.~Li, Z.~Liu, B.~Dai, T.~Zhao, and L.~Song, ``{Deep
  Hyperspherical Learning},'' in \emph{{Advances in Neural Information
  Processing Systems (NeurIPS)}}, 2017.

\bibitem{Liu2018CVPR}
W.~Liu, Z.~Liu, Z.~Yu, B.~Dai, R.~Lin, Y.~Wang, J.~M. Rehg, and L.~Song,
  ``{Decoupled Networks},'' in \emph{{Proceedings of the IEEE Conference on
  Computer Vision and Pattern Recognition (CVPR)}}, 2018.

\bibitem{luo2018cosine}
C.~Luo, J.~Zhan, X.~Xue, L.~Wang, R.~Ren, and Q.~Yang, ``{Cosine normalization:
  Using cosine similarity instead of dot product in neural networks},'' in
  \emph{{International Conference on Artificial Neural Networks (ICANN)}},
  2018.

\bibitem{ghiasi2019generalizing}
K.~Ghiasi-Shirazi, ``{Generalizing the convolution operator in convolutional
  neural networks},'' \emph{Neural Processing Letters}, vol.~50, no.~3, 2019.

\bibitem{wang2019kervolutional}
C.~Wang, J.~Yang, L.~Xie, and J.~Yuan, ``{Kervolutional neural networks},'' in
  \emph{{Proceedings of the IEEE Conference on Computer Vision and Pattern
  Recognition (CVPR)}}, 2019.

\bibitem{wang2019bias}
S.~Wang, T.~Zhou, and J.~Bilmes, ``{Bias also matters: Bias attribution for
  deep neural network explanation},'' in \emph{{International Conference on
  Machine Learning (ICML)}}, 2019.

\bibitem{mohan2019robust}
S.~Mohan, Z.~Kadkhodaie, E.~P. Simoncelli, and C.~Fernandez-Granda, ``{Robust
  And Interpretable Blind Image Denoising Via Bias-Free Convolutional Neural
  Networks},'' in \emph{{International Conference on Learning Representations
  (ICLR)}}, 2020.

\bibitem{hesse2021fast}
R.~Hesse, S.~Schaub-Meyer, and S.~Roth, ``{Fast Axiomatic Attribution for
  Neural Networks},'' in \emph{{Advances in Neural Information Processing
  Systems (NeurIPS)}}, vol.~34, 2021.

\bibitem{vaswani2017attention}
A.~Vaswani, N.~Shazeer, N.~Parmar, J.~Uszkoreit, L.~Jones, A.~N. Gomez,
  {\L}.~Kaiser, and I.~Polosukhin, ``{Attention is all you need},''
  \emph{{Advances in Neural Information Processing Systems (NeurIPS)}}, 2017.

\bibitem{serrano2019attention}
S.~Serrano and N.~A. Smith, ``{Is Attention Interpretable?}'' in
  \emph{{Proceedings of the Annual Meeting of the Association for Computational
  Linguistics (ACL)}}, 2019.

\bibitem{abnar2020quantifying}
S.~Abnar and W.~Zuidema, ``{Quantifying Attention Flow in Transformers},'' in
  \emph{{Proceedings of the Annual Meeting of the Association for Computational
  Linguistics (ACL)}}, 2020.

\bibitem{bastings2020}
J.~Bastings and K.~Filippova, ``{The elephant in the interpretability room: Why
  use attention as explanation when we have saliency methods?}'' in
  \emph{{Proceedings of the Third BlackboxNLP Workshop on Analyzing and
  Interpreting Neural Networks for NLP}}, 2020.

\bibitem{barkan2021GradSAM}
O.~Barkan, E.~Hauon, A.~Caciularu, O.~Katz, I.~Malkiel, O.~Armstrong, and
  N.~Koenigstein, ``{Grad-SAM: Explaining Transformers via Gradient
  Self-Attention Maps},'' in \emph{{Proceedings of the International Conference
  on Information and Knowledge Management (CIKM)}}, 2021.

\bibitem{chefer2021transformer}
H.~Chefer, S.~Gur, and L.~Wolf, ``{Transformer interpretability beyond
  attention visualization},'' in \emph{{Proceedings of the IEEE Conference on
  Computer Vision and Pattern Recognition (CVPR)}}, 2021.

\bibitem{swish}
P.~Ramachandran, B.~Zoph, and Q.~V. Le, ``{Searching for Activation
  Functions},'' in \emph{{International Conference on Learning Representations
  (ICLR)}, Workshop}, 2018.

\bibitem{goodfellow2013maxout}
I.~Goodfellow, D.~Warde-Farley, M.~Mirza, A.~Courville, and Y.~Bengio,
  ``{Maxout networks},'' in \emph{{International Conference on Machine Learning
  (ICML)}}, 2013.

\bibitem{ioffe2015batchnorm}
S.~Ioffe and C.~Szegedy, ``{Batch Normalization: Accelerating Deep Network
  Training by Reducing Internal Covariate Shift},'' in \emph{{International
  Conference on Machine Learning (ICML)}}, 2015.

\bibitem{ba2016layer}
J.~L. Ba, J.~R. Kiros, and G.~E. Hinton, ``{Layer normalization},''
  \emph{CoRR}, vol. abs/1607.06450, 2016.

\bibitem{li2019positional}
B.~Li, F.~Wu, K.~Q. Weinberger, and S.~Belongie, ``{Positional
  normalization},'' \emph{Advances in Neural Information Processing Systems},
  vol.~32, 2019.

\bibitem{ulyanov2016instance}
D.~Ulyanov, A.~Vedaldi, and V.~Lempitsky, ``{Instance normalization: The
  missing ingredient for fast stylization},'' \emph{CoRR}, vol. abs/1607.08022,
  2016.

\bibitem{wu2018group}
Y.~Wu and K.~He, ``{Group normalization},'' in \emph{{Proceedings of the
  European conference on computer vision (ECCV)}}, 2018.

\bibitem{deng2009imagenet}
J.~Deng, W.~Dong, R.~Socher, L.~Li, K.~Li, and L.~Fei{-}Fei, ``{ImageNet: {A}
  large-scale hierarchical image database},'' in \emph{{Proceedings of the IEEE
  Conference on Computer Vision and Pattern Recognition (CVPR)}}, 2009.

\bibitem{pytorch}
A.~Paszke, S.~Gross, F.~Massa, A.~Lerer, J.~Bradbury, G.~Chanan, T.~Killeen,
  Z.~Lin, N.~Gimelshein, L.~Antiga, A.~Desmaison, A.~Kopf, E.~Yang, Z.~DeVito,
  M.~Raison, A.~Tejani, S.~Chilamkurthy, B.~Steiner, L.~Fang, J.~Bai, and
  S.~Chintala, ``{PyTorch: An Imperative Style, High-Performance Deep Learning
  Library},'' in \emph{{Advances in Neural Information Processing Systems
  (NeurIPS)}}, 2019.

\bibitem{beyer2022better}
L.~Beyer, X.~Zhai, and A.~Kolesnikov, ``{Better plain ViT baselines for
  ImageNet-1k},'' \emph{CoRR}, vol. abs/2205.01580, 2022.

\bibitem{xiao2021early}
T.~Xiao, M.~Singh, E.~Mintun, T.~Darrell, P.~Doll{\'a}r, and R.~Girshick,
  ``{Early convolutions help transformers see better},'' \emph{Advances in
  Neural Information Processing Systems (NeurIPS)}, vol.~34, 2021.

\bibitem{torchvision}
``{Torchvision library, pretrained models},''
  \url{https://pytorch.org/vision/stable/models.html}, accessed: 2021-11-11.

\bibitem{torchvisionConvnext}
``{How to Train State-Of-The-Art Models Using TorchVision’s Latest
  Primitives},''
  \url{https://pytorch.org/blog/how-to-train-state-of-the-art-models-using-torchvision-latest-primitives/},
  2021, accessed: 2023-05-23.

\bibitem{baehrens2010explain}
D.~Baehrens, T.~Schroeter, S.~Harmeling, M.~Kawanabe, K.~Hansen, and K.-R.
  M{\"u}ller, ``{How to explain individual classification decisions},''
  \emph{The Journal of Machine Learning Research (JMLR)}, 2010.

\bibitem{rao2022towards}
S.~Rao, M.~B{\"o}hle, and B.~Schiele, ``{Towards better understanding
  attribution methods},'' in \emph{{Proceedings of the IEEE/CVF Conference on
  Computer Vision and Pattern Recognition (CVPR)}}, 2022.

\end{thebibliography}

\begin{IEEEbiography}[{\includegraphics[width=1in,height=1.25in,clip,keepaspectratio]{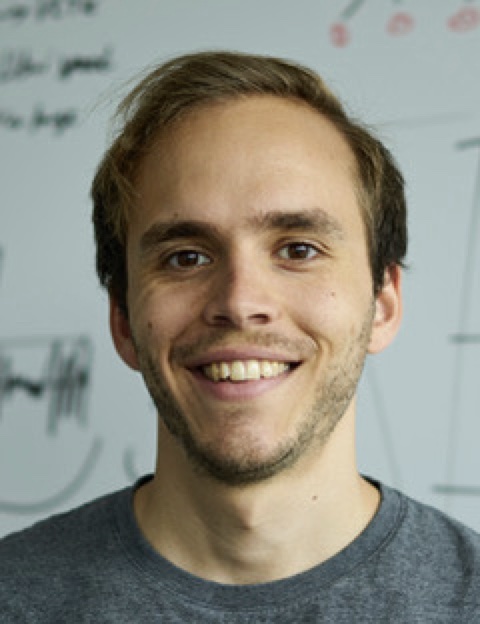}}]{Moritz Böhle}
is a Ph.D. student in Computer Science at the Max Planck Institute for Informatics, working with Prof. Dr. Bernt Schiele and Prof. Dr. Mario Fritz. He graduated with a bachelor's degree in physics in 2016 from the Freie Universität Berlin and obtained his Master’s degree in computational neuroscience in 2019 from the Technische Universität Berlin. His research focuses on understanding the `decision process' in deep neural networks and designing inherently interpretable neural network models.
\end{IEEEbiography}

\begin{IEEEbiography}[{\includegraphics[width=1in,height=1.25in,clip,keepaspectratio]{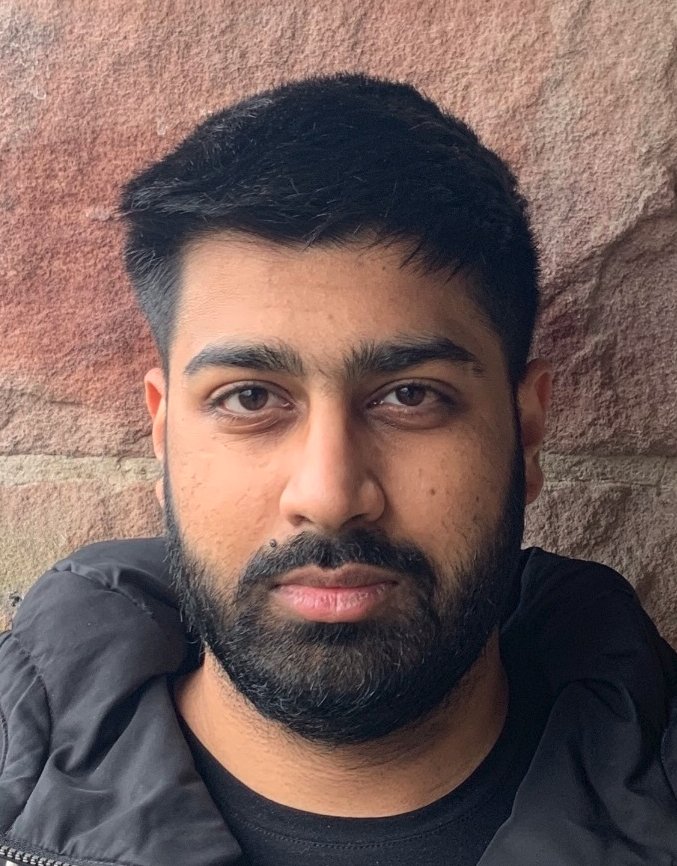}}]{Navdeeppal Singh}
is a Master's student pursuing Computer Science at Saarland University, where he also obtained his Bachelors's degree in Computer Science in 2021. 
For his bachelor thesis, he worked on the robustness evaluation of inherently interpretable neural network models. He's interested in computer vision, interpretable and robust neural networks, and neural search.
\end{IEEEbiography}

\begin{IEEEbiography}[{\includegraphics[width=1in, height=1.25in, trim=6em 0 6em 0, clip, keepaspectratio]{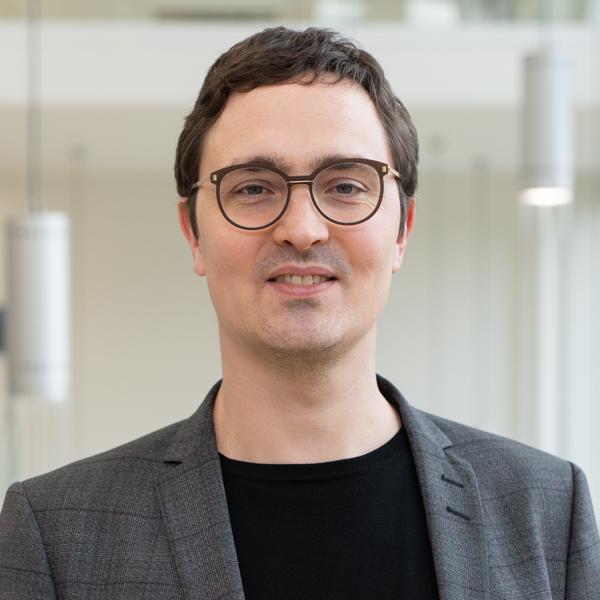}}]{Mario Fritz}
Mario Fritz is faculty member at the CISPA Helmholtz Center for Information Security, honorary professor at the Saarland University, and a fellow of the European Laboratory of Learning and Intelligent Systems (ELLIS). Before, he was senior researcher at the Max Planck Institute for Informatics, PostDoc at UC Berkeley and International Computer Science Institute. He studied computer science at the university Erlangen/Nuremberg and obtained his PhD from TU Darmstadt. His current work is centered around Trustworthy Information Processing with a focus on the intersection of AI \& Machine Learning with Security \& Privacy. He is associate editor of IEEE TPAMI, and a leading scientist of the Helmholtz Medical Security, Privacy, and AI Research Center, where he is coordinating projects on privacy and federated learning in health. He has over 100 publications, including 80 in top-tier journals (IJCV, TPAMI) and conferences (NeurIPS, AAAI, IJCAI, ICLR, NDSS, USENIX Security, CCS, S\&P, CVPR, ICCV, ECCV).
\end{IEEEbiography}

\begin{IEEEbiography}[{\includegraphics[width=1in,height=1.25in,clip,keepaspectratio]{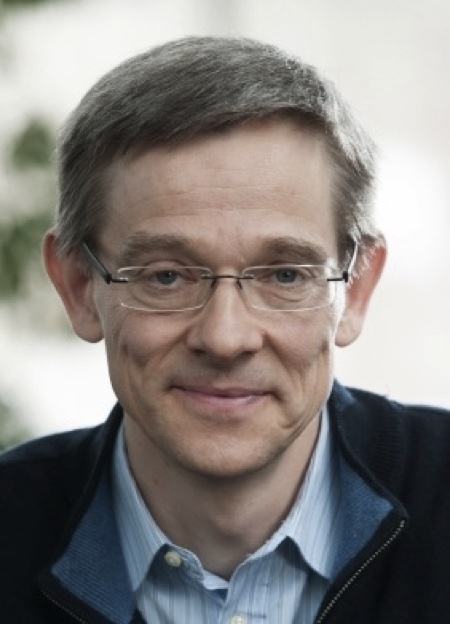}}]{Bernt Schiele}
has been Max Planck Director at MPI for Informatics and Professor at Saarland University since 2010. He studied computer science at the University of Karlsruhe, Germany. He worked on his master thesis in the field of robotics in Grenoble, France, where he also obtained the ``diplome d'etudes approfondies d'informatique''. In 1994 he worked in the field of multi-modal human-computer interfaces at Carnegie Mellon University, Pittsburgh, PA, USA in the group of Alex Waibel. In 1997 he obtained his PhD from INP Grenoble, France under the supervision of Prof. James L. Crowley in the field of computer vision. The title of his thesis was “Object Recognition using Multidimensional Receptive Field Histograms”. Between 1997 and 2000 he was postdoctoral associate and Visiting Assistant Professor with the group of Prof. Alex Pentland at the Media Laboratory of the Massachusetts Institute of Technology, Cambridge, MA, USA. From 1999 until 2004 he was Assistant Professor at the Swiss Federal Institute of Technology in Zurich (ETH Zurich). Between 2004 and 2010 he was Full Professor at the computer science department of TU Darmstadt. He is fellow of IEEE, ACM, ELLIS, and IAPR. 
\end{IEEEbiography}

\clearpage


\section*{Supplementary material}

# B-cos Alignment for Inherently Interpretable CNNs and Vision Transformers

## Supplementary Material

Moritz Böhle
MPI for Informatics
Saarland Informatics Campus

Navdeeppal Singh
MPI for Informatics
Saarland Informatics Campus

Mario Fritz
CISPA Helmholtz Center
for Information Security

Bernt Schiele
MPI for Informatics
Saarland Informatics Campus

## Table of Contents

In this supplement to our work on B-cos DNNs, we provide:

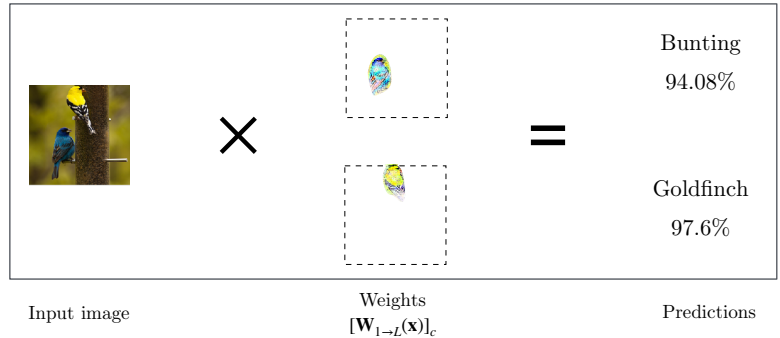

Fig. A1. Illustration of the computations of a B-cos network. For a given input image (left), the model computes an *input-dependent* linear transform $\mathbf{W}_{1\to L}(\mathbf{x})$ (center). The scalar product between the input and the weights $[\mathbf{W}_{1\to L}(\mathbf{x})]_c$ for class $c$ (row $c$ of $\mathbf{W}_{1\to L}(\mathbf{x})$), yields the class logits for the respective class. To obtain class probabilities (right), we apply the sigmoid function. Since the B-cos networks are trained with the BCE loss, they produce probabilities *per class* and *not a probability distribution over classes*. Thus, the probabilities do not sum to 1. For illustration purposes, we only visualise the positive contributions according to $\mathbf{W}_{1\to L}(\mathbf{x})$.

## A. Additional qualitative examples

In Fig. A1, we illustrate how the linear mappings $\mathbf{W}_{1\to L}(\mathbf{x})$ are used to compute the outputs of B-cos networks. In particular, with this we would like to highlight that these linear mappings do not only constitute qualitatively convincing visualisations. Instead, they are in fact based on the actual linear transformation matrix that the model effectively applies to the input to compute its outputs and thus constitute an accurate summary of the model computations.

**Comparison between model explanations.** In Figs. A2 and A3, we compare explanations for different network architec-

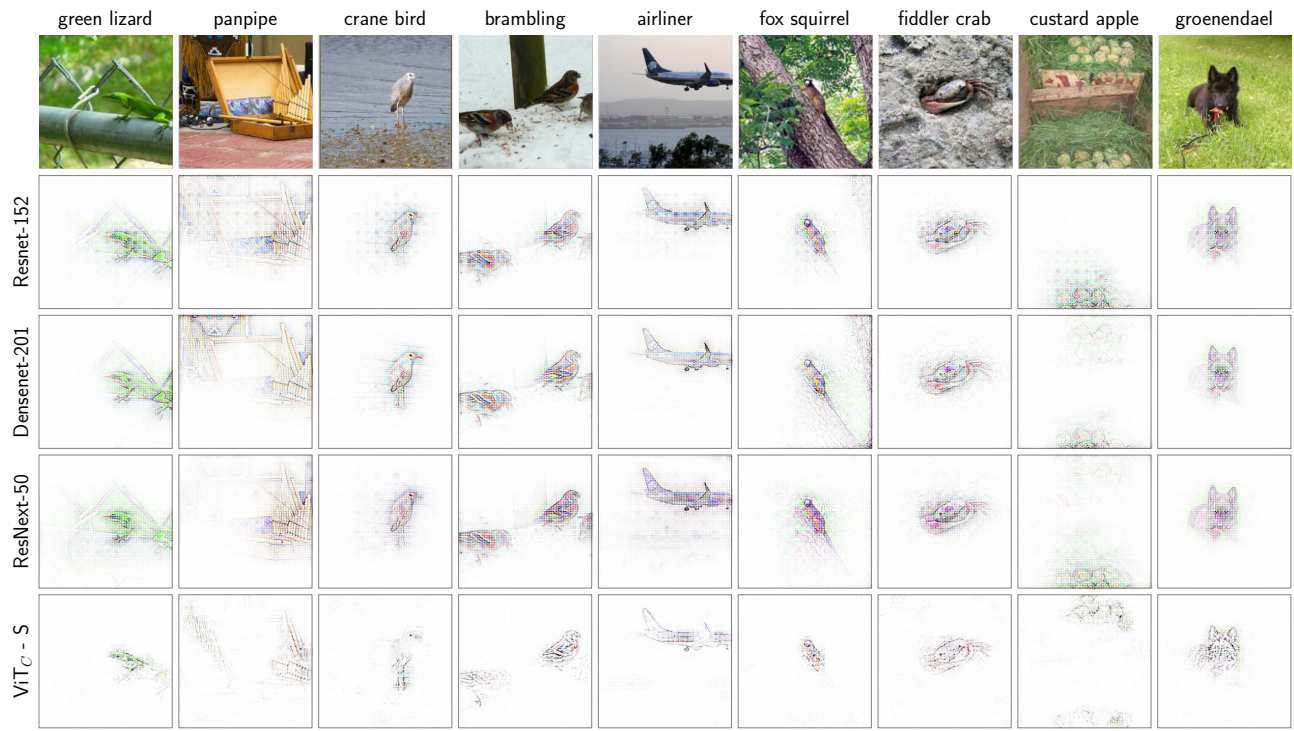

Fig. A2. **Additional model explanations** based on $\mathbf{W}(\mathbf{x})$ for various B-cos architectures, see also Fig. A3. While the explanations are generally consistent between models, architecture-specific details can be observed. E.g., ResNet-based architectures exhibit a grid-pattern in the explanations, and we find that the ViT models generally produce sparser explanations.

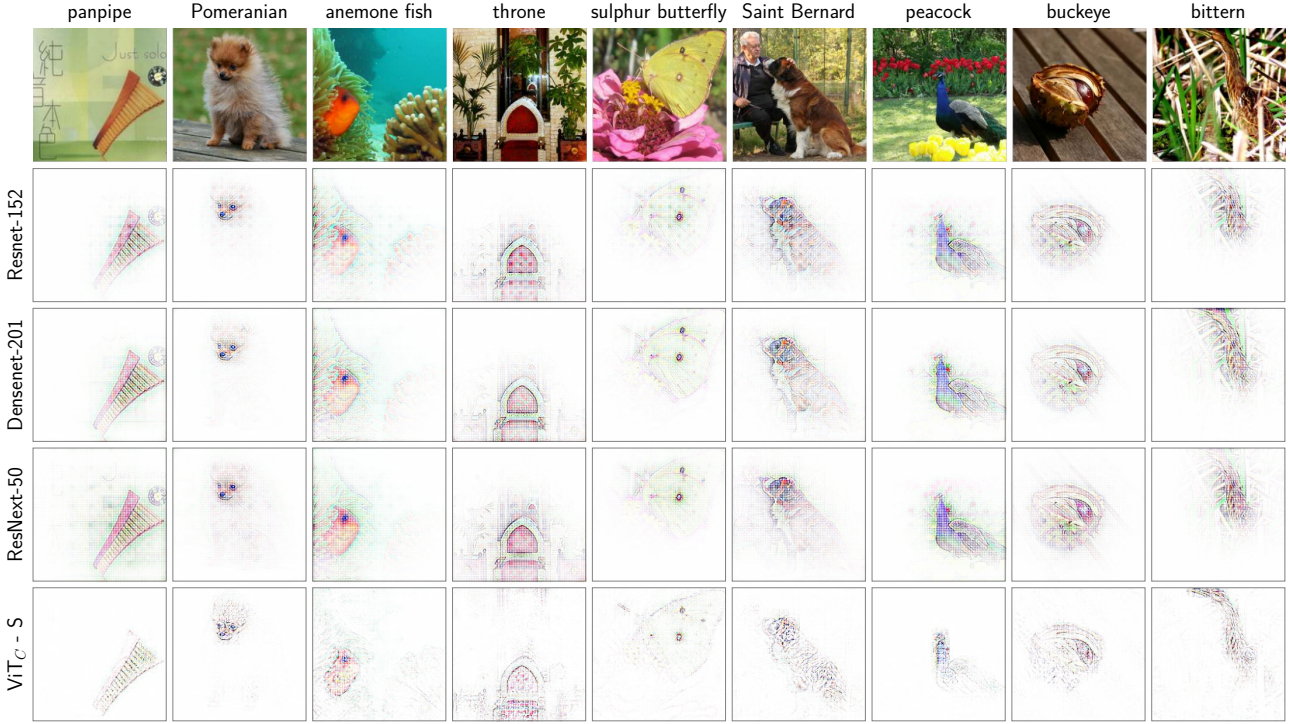

Fig. A3. Model explanations based on $\mathbf{W}(\mathbf{x})$ for various B-cos architectures, see also Fig. A2. While the explanations are generally consistent between models, architecture-specific details can be observed. E.g., ResNet-based architectures exhibit a grid-pattern in the explanations, and we find that the ViT models generally produce sparser explanations.

tures on the same set of images. We find that all models generally seem to focus on the same regions in the image. However, we also observe architecture-specific details: *e.g.*, we find that the vision transformer yield much sparser explanations than the convolutional models and ResNet models (ResNet-152 and ResNext-50 in the figures) exhibit grid-pattern artefacts, which we attribute to the downsampling mechanism in those models. Considering that information is not processed as locally in the ViT$_C$ models as in the convolutional models, these qualitative results provide further evidence for the model-faithfulness of the explanations based on $\mathbf{W}(\mathbf{x})$.

## B. Additional quantitative evaluations

In Fig. B1, we present the localisation results of the grid pointing game [S5] for two additional model architectures, see also Fig. 6 in the main paper.

In particular, in Fig. B1 (left), we show that of all methods, the best explanation for the B-cos models is given by the model-inherent linear transformations $\mathbf{W}_{1 \to L}(\mathbf{x})$. Moreover, in Fig. B1 (right) we can see that the B-cos explanations yield a significant interpretability gain over the respective baselines: specifically, we see that no method explains the baseline models better than the model-inherent linear transformations explain the respective B-cos network.

Finally, as discussed in the main paper (see Fig. 7), the results of the B-cos explanations can be further improved by including only the top-n pixels (here ≈11.3k) in the explanations, which we denote by OursQ in the figure.

## C. Implementation details

In the following, we provide further implementation details regarding implementation of a convolutional B-cos transform (Algorithm 1), the training and evaluation procedure (C.1), and the post-hoc attribution methods (C.2).

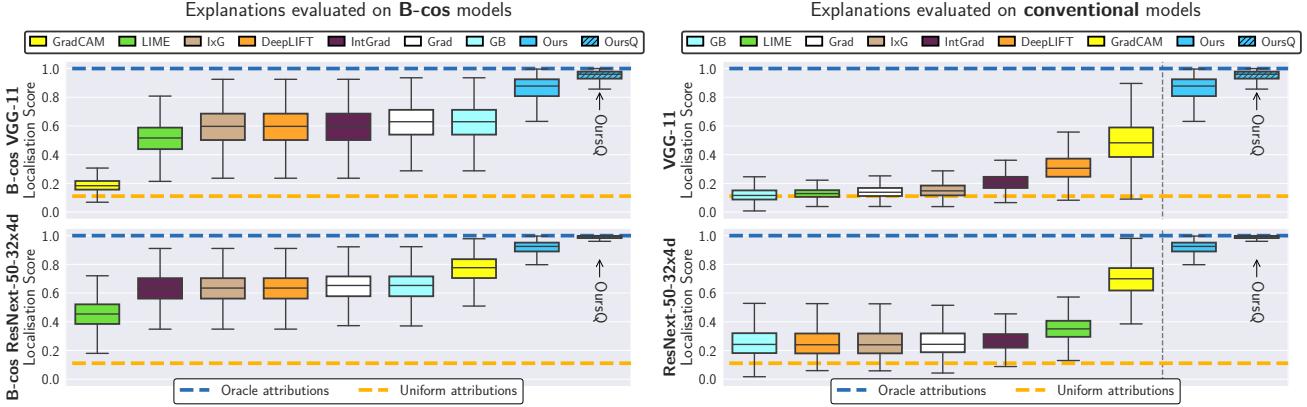

Fig. B1. **Additional localisation results.** We show localisation results of two additional B-cos models (left) and the conventional counterparts (right). These results are consistent with those shown in Fig. 6 in the main paper. Specifically, we find the model-inherent explanations (Ours) of the B-cos models to yield close to optimal localisation results, especially when only including the top 11.3k pixels in the explanations (OursQ). Additionally, and in accordance with the results reported in the main paper for a DenseNet and a ResNet model, we find that the B-cos models are better explained by post-hoc explanation methods than the conventional models, except for GradCAM on VGG. For VGG, the gradient computation of GradCAM seems to be significantly affected by the non-linear aspects of the B-cos transformations (note that VGG is the only model using a multi-layer perceptron for the classification head).

---

**Algorithm 1:** Pseudocode for B-cos-Conv2d, cf. Eq. (10) in the main paper.

---

1 # **x**: input, $\widehat{\mathbf{W}}$: normed weights, $k$: kernel size, $d_f$: index of feature dimension
2 **def** bcos_conv2d(**x**, $\widehat{\mathbf{W}}$, $k$, $B$):
3      linear_out = conv2d(**x**, $\widehat{\mathbf{W}}$) # $= \widehat{\mathbf{W}}\mathbf{x}$
4      norm = sumpool2d(**x**.pow(2).sum($d_f$), k).sqrt()
5      cos = linear_out / norm.unsqueeze($d_f$)
6      scaling = cos.abs().pow(B-1) # $= |c(\mathbf{x}; \widehat{\mathbf{W}})|^{B-1}$
7      **return** scaling * linear_out # $= |c(\mathbf{x}; \widehat{\mathbf{W}})|^{B-1}\widehat{\mathbf{W}}\mathbf{x}$

## C.1. Training and evaluation procedure

### C.1.1 CIFAR10

**Architecture.** For our simple B-cos models trained on CIFAR10, we used a 9-layer architecture with the following specifications: kernel size $k = [3, 3, 3, 3, 3, 3, 3, 3, 1]$, stride $s = [1, 1, 2, 1, 1, 2, 1, 1, 1]$, padding $p = [1, 1, 1, 1, 1, 1, 1, 1, 0]$, and output channels $o = [64, 64, 128, 128, 128, 256, 256, 256, 10]$ for layers $l = [1, 2, 3, 4, 5, 6, 7, 8, 9]$ respectively.

When increasing the parameter B, we observed the input signal to decay strongly over the network layers, which resulted in zero outputs and hindered training. To overcome this, we scaled all layer outputs with a fixed scalar $\gamma$ to improve signal propagation, which we set such that $\log_{10}\gamma = 1.75 + B/10$.. To counteract the artificial upscaling of the signal at the network output, we down-scaled the network output by a fixed constant $T$ for each B, such that $\log_{10} T = [8.9, 8.125, 7.35, 6.757, 4.8, 4.525, 4.25]$ for B $= [1, 1.25, 1.5, 1.75, 2, 2.25, 2.5]$ respectively.

Furthermore, for our B-cos ResNet models for CIFAR10 we converted the ResNet-{20,32,44,56} [S7] models to their corresponding B-cos version. Specifically, to evaluate the applicability of different normalisation layers, we replaced the conventionally used Batch-Norm layers by Batch-, Layer-, Instance-, Position-, and All-Norm layers (Eqs. (20)–(24)). As we report in Fig. 13 in the main paper, for some of the normalisation layers (*e.g.* All-, Instance-, and Batch-Norm), the bias terms can play a significant role in the classification decision. Therefore, for the ImageNet experiments, we exclusively use *bias-free* normalisation layers. Furthermore, we switch the order of the the global pooling layer and last linear classifier layer, *i.e.* we first do a linear classification (using a convolution) before doing a global (average) pooling, in order to more easily evaluate those models in the localisation metric, for which the size of the input image changes. Note that for models with a linear classification head, this yields functionally equivalent models; for the B-cos models, however, the ordering matters,

4

since the classification head is a non-linear (*i.e.* B-cos) layer.

**Training.** We trained all our CIFAR10 models for 100 epochs with the Adam optimiser [S9], an initial learning rate of $1 \times 10^{-3}$, and a batch size of 64. Further, we used a cosine learning rate schedule and decayed the learning rate to $1 \times 10^{-5}$ for our simple B-cos models and to 0 for our B-cos ResNets over 100 epochs. For augmenting the data, we applied horizontal flipping and padded random cropping. We used a bias term of $\mathbf{b} = \log(0.1/0.9)$, which yields a uniform probability distribution for zero inputs ($[\mathbf{f}(\mathbf{x} = \mathbf{0})]_i = [\sigma(\mathbf{W}_{1 \to L} \mathbf{0} + \mathbf{b})]_i = 0.1 \ \forall \ i$).

### C.1.2 ImageNet

**Training.** Our training recipe is based on the `torchvision` (https://github.com/pytorch/vision/) reference recipes. For our ResNets, DenseNets, and VGGs, we train for 90 epochs with the Adam optimiser using a learning rate of $1 \times 10-3$ and a batch size of 256 (distributed over 4 GPUs with each having batch size of 64). Following `torchvision`, we also train our ResNeXt model with the same recipe as for the ResNets but for 100 epochs instead of 90. As mentioned in Sec. 3.2.2 in the main paper, we change the target encoding $\mathbf{y}_i$ for non-target classes to be $1/C$. For the learning rate schedule, we first use a linear warm-up for 5 epochs, then we use a cosine annealing learning schedule, decaying the learning rate to 0. To avoid divergence issues we employ gradient clipping. However, we empirically noticed that the magnitude of the gradient for B-cos models tends to be significantly lower than that of the gradients for standard networks. To avoid manually tuning the clipping threshold, we opt for employing Adaptive Gradient Clipping (AGC) [S4]. For data augmentation, we use random cropping and then resizing with bilinear interpolation to size 224 along with random horizontal flips with a probability of 0.5.

Inspired by `torchvision`'s new longer training recipes[1], we additionally train a set of ConvNeXt-{Tiny, Base}, DenseNet121, and ResNet{50, 152} models for 600 epochs with the Adam optimiser using a learning rate of $1 \times 10^{-3}$ and a batch size of 1024 (distributed over 8 GPUs with each having a batch size of 128). We keep an exponential moving average (EMA) of the weights of the model with a decay rate of 0.99998 updated after every 32 training steps. We use the same learning rate scheduler as mentioned above and AGC for gradient clipping. For data augmentation, we use random cropping and then resizing to size 176 with bilinear interpolation, MixUp [S17] with $\alpha = 0.2$, CutMix [S16] with $\alpha = 1.0$, Repeated Augmentations [S2, S8] with 4 repetitions, Random Erasing [S6, S18] with a probability of 0.1, horizontal flipping with a probability of 0.5, and finally TrivialAugment [S10] which is parameter-free.

We trained our ViTs from scratch following the 90 epoch protocol described by [S3]. Given that all the B-cos layer weights are normalised to unit norm, we employ Adam instead of AdamW, since no weight decay is necessary. Furthermore, as mentioned previously we employ AGC over standard gradient clipping. We used an effective batch size of 1024 (distributed over 8 GPUs) for all but our largest ViTs (ViT-L models and the Simple ViT-B model), for which we use a batch size of 512 due to memory constraints. We observed these larger ViT models to diverge at the beginning of training and therefore opted to increase the length of the warm-up period from the 10 000 steps described by [S3] to 50 000 steps.

For all further details, we kindly refer the reader to the code at https://github.com/B-cos/B-cos-v2, which we make available along with this submission.

**Evaluation.** We evaluated all networks on the full images of the ImageNet validation set, after rescaling the smaller dimension (height / width) to 256 using bilinear interpolation. For networks with EMA weights, we evaluate both the EMA weights non-EMA weights and report the best performing one.

### C.2. Attribution methods

We compare the model-inherent explanations, given by the linear transformation $\mathbf{W}_{1 \to L}(\mathbf{x})$, against the following post-hoc attribution methods: the vanilla gradient (Grad, [S1]), 'Input×Gradient' (IxG, [S13]), Guided Backpropagation' (GB, [S14]), Integrated Gradients (IntGrad, [S15]), DeepLIFT ( [S13]), GradCAM ( [S12]), and LIME ( [S11]).

For all methods except LIME, we rely on the captum library (github.com/pytorch/captum). For IntGrad, we set n_steps = 50 for integrating over the gradients. For LIME, we used the official implementation available at github.com/marotcr/lime, with 500 samples and the default values for the kernel size ($k = 4$) and evaluate the weights assigned to the individual image segments, see Fig. 9 in the main paper for a visualisation.

### C.2.1 Localisation metric

We evaluated all attribution methods on the grid pointing game [S5]. For this, we constructed 500 $3 \times 3$ grid images. For an example of a $2 \times 2$ grid, see Fig. 3 in the main paper. As was done in [S5], we sorted the images according to the

---
[1] https://pytorch.org/blog/how-to-train-state-of-the-art-models-using-torchvision-latest-primitives/

models' classification confidence for each class and then sampled a random set of classes for each multi-image. For each of the sampled classes, we then included the most confidently classified image in the grid that had not already been used in a previous grid image.

## D. Additional derivations and discussions

### D.1. The B-cos transformation as a scaled linear transformation

In the following, we provide additional details on how to express the B-cos transformation as a scaled linear transformation and in matrix form.

As described in Eq. (3) in the main paper, the B-cos transformation is given by

$$\text{B-cos}(\mathbf{x}; \mathbf{w}) = \|\widehat{\mathbf{w}}\| \, \|\mathbf{x}\| \times |c(\mathbf{x}, \widehat{\mathbf{w}})|^{\text{B}} \times \text{sgn}(c(\mathbf{x}, \widehat{\mathbf{w}})) \,, \tag{D.1}$$

$$\text{with} \qquad c(\mathbf{x}, \mathbf{w}) = \cos(\angle(\mathbf{x}, \mathbf{w})) \,, \tag{D.2}$$

$$\widehat{\mathbf{w}} = \mathbf{w}/\|\mathbf{w}\| \,, \tag{D.3}$$

$\angle(\mathbf{x}, \mathbf{w})$ returning the angle between $\mathbf{x}$ and $\mathbf{w}$, and sgn the sign function. Note that the sign function can be expressed as $\text{sgn}(a) = a/|a|$ for $|a| \neq 0$ and zero otherwise. Hence, Eq. (D.1) can be expressed as

$$\text{B-cos}(\mathbf{x}; \mathbf{w}) = \|\widehat{\mathbf{w}}\| \, \|\mathbf{x}\| \times |c(\mathbf{x}, \widehat{\mathbf{w}})|^{\text{B}} \times \text{sgn}(c(\mathbf{x}, \widehat{\mathbf{w}})) \tag{D.4}$$

$$\text{(replace sgn)} \qquad = \|\widehat{\mathbf{w}}\| \, \|\mathbf{x}\| \times |c(\mathbf{x}, \widehat{\mathbf{w}})|^{\text{B}} \times c(\mathbf{x}, \widehat{\mathbf{w}})/|c(\mathbf{x}, \widehat{\mathbf{w}})| \tag{D.5}$$

$$\text{(combine cos terms)} \qquad = \|\widehat{\mathbf{w}}\| \, \|\mathbf{x}\| \times |c(\mathbf{x}, \widehat{\mathbf{w}})|^{\text{B}-1} \times c(\mathbf{x}, \widehat{\mathbf{w}}) \tag{D.6}$$

$$\text{(reorder)} \qquad = \|\widehat{\mathbf{w}}\| \, \|\mathbf{x}\| \times c(\mathbf{x}, \widehat{\mathbf{w}}) \times |c(\mathbf{x}, \widehat{\mathbf{w}})|^{\text{B}-1} \tag{D.7}$$

$$\text{(write first three factors as linear transform)} \qquad = \widehat{\mathbf{w}}^T \mathbf{x} \times |c(\mathbf{x}, \widehat{\mathbf{w}})|^{\text{B}-1} \,. \tag{D.8}$$

For clarity, we marked the changes between lines in the above equations in red.

From Eq. (D.8) it becomes clear that a B-cos transformation simply computes a rescaled linear transform, as in Eq. (4) in the main paper. Thus, multiple units in parallel (i.e., a layer $\mathbf{l}^*$ of B-cos units) can easily be expressed in matrix form via

$$\mathbf{l}^*(\mathbf{x}) = |c(\mathbf{x}, \widehat{\mathbf{W}})|^{\text{B}-1} \times \widehat{\mathbf{W}} \mathbf{x} \,. \tag{D.9}$$

Here, the $\times$, cos, and absolute value operators are applied element-wise and the rows of $\widehat{\mathbf{W}}$ are given by $\widehat{\mathbf{w}}_n$ of the individual units $n$. Hence, the output of each unit (entry in output vector $\mathbf{l}^*$) is the down-scaled linear transformation from Eq. (D.8). Note that Eq. (D.9) is the same as Eq. (10) in the main paper.

### D.2. On the relevance of image encoding for the visualisations

As we describe in the main paper, we encode image pixels as $[r, g, b, 1-r, 1-g, 1-b]$. This has two important advantages.

On the one hand, as argued by [S5], this overcomes a bias towards bright pixels. For this, note that the model output is computed as a linear transformation of the input $\mathbf{x}$. As such, the contribution to the output per pixel is given by the weighted input strength. In particular, a specific pixel location $(i, j)$ with color channels $c$ contributes $\sum_c w_{(i,j,c)} x_{(i,j,c)}$ to the output. Under the conventional encoding—i.e., $[r, g, b]$—, a black pixel is encoded by $x_{(i,j,c)} = 0$ for $c \in \{1, 2, 3\}$ and can therefore not contribute to the model output. Since we train the model to maximise its outputs (binary cross entropy loss, see Sec. 3.2.2 in the main paper), the network will preferentially encode bright pixels, as these can produce higher contributions for maximising the output than dark pixels. In contrast, under the new encoding dark and bright pixels have the same amount of 'signal' that can be weighted, i.e., $\sum_c x_{(i,j,c)} = 3 \; \forall \; (i, j)$.

Moreover, this encoding allows to unambiguously infer the color of a pixel solely based on the angle of the pixel vector $[r, g, b, 1-r, 1-g, 1-b]$. To contrast this with the original encoding, consider a pixel that is (almost) completely black and given by $[r, g, b]$ with $g = 0, b = 0, r = 0.001$. This pixel has the same angle as a red pixel, given by $r = 1, g = 0, b = 0$. Thus, these two colors cannot be disambiguated based on their angle. By adding the three additional color channels $[1-r, 1-g, 1-b]$, each color is uniquely encoded by the direction of the pixel vector. Finally, note that the B-cos transformation induces an alignment pressure on the weights, i.e., the model weights are optimised such that $\mathbf{W}_{1 \rightarrow L}$ points in the same direction as (important features in) the input. Consequently, the weights will reproduce the *angles* of the pixels, but there is no constraint on their *norm*. Since the angle is sufficient for inferring the color, we can nevertheless decode the

angles of the weight vectors into RGB colors.

## References

[S1] David Baehrens, Timon Schroeter, Stefan Harmeling, Motoaki Kawanabe, Katja Hansen, and Klaus-Robert Müller. How to explain individual classification decisions. *The Journal of Machine Learning Research (JMLR)*, 2010. 5

[S2] Maxim Berman, Hervé Jégou, Andrea Vedaldi, Iasonas Kokkinos, and Matthijs Douze. MultiGrain: a unified image embedding for classes and instances. *CoRR*, abs/1902.05509, 2019. 5

[S3] Lucas Beyer, Xiaohua Zhai, and Alexander Kolesnikov. Better plain ViT baselines for ImageNet-1k. *CoRR*, abs/2205.01580, 2022. 5

[S4] Andy Brock, Soham De, Samuel L Smith, and Karen Simonyan. High-Performance Large-Scale Image Recognition Without Normalization. In *International Conference on Machine Learning (ICML)*, 2021. 5

[S5] Moritz Böhle, Mario Fritz, and Bernt Schiele. Convolutional Dynamic Alignment Networks for Interpretable Classifications. In *Proceedings of the IEEE Conference on Computer Vision and Pattern Recognition (CVPR)*, 2021. 3, 5, 6

[S6] Terrance Devries and Graham W. Taylor. Improved Regularization of Convolutional Neural Networks with Cutout. *CoRR*, abs/1708.04552, 2017. 5

[S7] Kaiming He, Xiangyu Zhang, Shaoqing Ren, and Jian Sun. Deep residual learning for image recognition. In *Proceedings of the IEEE Conference on Computer Vision and Pattern Recognition (CVPR)*, 2016. 4

[S8] Elad Hoffer, Tal Ben-Nun, Itay Hubara, Niv Giladi, Torsten Hoefler, and Daniel Soudry. Augment Your Batch: Improving Generalization Through Instance Repetition. In *Proceedings of the IEEE Conference on Computer Vision and Pattern Recognition (CVPR)*, 2020. 5

[S9] Diederik P. Kingma and Jimmy Ba. Adam: A Method for Stochastic Optimization. In *International Conference on Learning Representations (ICLR)*, 2015. 5

[S10] Samuel G. Müller and Frank Hutter. TrivialAugment: Tuning-free Yet State-of-the-Art Data Augmentation. In *International Conference on Computer Vision (ICCV)*, 2021. 5

[S11] Marco Tulio Ribeiro, Sameer Singh, and Carlos Guestrin. "Why Should I Trust You?": Explaining the Predictions of Any Classifier. In *International Conference on Knowledge Discovery and Data Mining (SIGKDD)*, 2016. 5

[S12] Ramprasaath R. Selvaraju, Michael Cogswell, Abhishek Das, Ramakrishna Vedantam, Devi Parikh, and Dhruv Batra. Grad-CAM: Visual Explanations from Deep Networks via Gradient-Based Localization. In *International Conference on Computer Vision (ICCV)*, 2017. 5

[S13] Avanti Shrikumar, Peyton Greenside, and Anshul Kundaje. Learning Important Features Through Propagating Activation Differences. In *International Conference on Machine Learning (ICML)*, 2017. 5

[S14] Jost Tobias Springenberg, Alexey Dosovitskiy, Thomas Brox, and Martin A. Riedmiller. Striving for Simplicity: The All Convolutional Net. In *International Conference on Learning Representations (ICLR), Workshop*, 2015. 5

[S15] Mukund Sundararajan, Ankur Taly, and Qiqi Yan. Axiomatic Attribution for Deep Networks. In *International Conference on Machine Learning (ICML)*, 2017. 5

[S16] Sangdoo Yun, Dongyoon Han, Seong Joon Oh, Sanghyuk Chun, Junsuk Choe, and Youngjoon 5 Yoo. CutMix: Regularization Strategy to Train Strong Classifiers with Localizable Features. In *International Conference on Computer Vision (ICCV)*, 2019. 5

[S17] Hongyi Zhang, Moustapha Cissé, Yann N. Dauphin, and David Lopez-Paz. mixup: Beyond Empirical Risk Minimization. In *International Conference on Learning Representations (ICLR)*, 2018. 5

[S18] Zhun Zhong, Liang Zheng, Guoliang Kang, Shaozi Li, and Yi Yang. Random Erasing Data Augmentation. In *The Thirty-Fourth AAAI Conference on Artificial Intelligence AAAI*, 2020. 5


\end{document}